\pdfoutput=1

\documentclass[11pt]{article}

\usepackage[]{EMNLP2022}

\usepackage{times}
\usepackage{latexsym}
\usepackage{xspace}

\usepackage[T1]{fontenc}

\usepackage[utf8]{inputenc}

\usepackage{microtype}

\usepackage{inconsolata}

\usepackage[russian,english]{babel}

\usepackage{amsmath}
\usepackage{graphicx}
\usepackage{multirow}
\usepackage{booktabs}
\usepackage{cuted}

\usepackage{tcolorbox}
\usepackage{array}  
\usepackage[caption=false]{subfig}

\usepackage{CJKutf8}        

\newcommand{\Sref}[1]{\S\ref{#1}}

\newcommand{\Fref}[1]{Figure~\ref{#1}}

\newcommand{\Tref}[1]{Table~\ref{#1}}
\newcommand{\Aref}[1]{Appendix~\ref{#1}}

\newcommand{\dataname}{VoynaSlov\xspace}
\newcommand{\datanamevk}{VoynaSlov-VK\xspace}
\newcommand{\datanametwitter}{VoynaSlov-Twitter\xspace}

\newcommand{\russian}[1]{\selectlanguage{russian}#1}

\title{Challenges and Opportunities in Information Manipulation Detection:\\ An Examination of Wartime Russian Media}
\date{}

\author{Chan Young Park\textsuperscript{*} \\
  Carnegie Mellon University \\
  \texttt{chanyoun@cs.cmu.edu} \\\And
  Julia Mendelsohn\textsuperscript{*} \\
  University of Michigan \\
  \texttt{juliame@umich.edu} \\\AND
  Anjalie Field\textsuperscript{*} \\
  Stanford University \\
  \texttt{anjalief@stanford.edu} \\\And
  Yulia Tsvetkov \\
  University of Washington \\
  \texttt{yuliats@cs.washington.edu} \\}

\begin{document}
\maketitle
\def\thefootnote{*}\footnotetext{Equal contribution}\def\thefootnote{\arabic{footnote}}

\begin{abstract}
NLP research on public opinion manipulation campaigns has primarily focused on detecting overt strategies such as fake news and disinformation. 
However, information manipulation in the ongoing Russia-Ukraine war exemplifies how governments and media also employ more nuanced strategies. 
We release a new dataset, \dataname{}, containing 38M+ posts from Russian media outlets on Twitter and VKontakte, as well as public activity and responses, immediately preceding and during the 2022 Russia-Ukraine war. We apply standard and recently-developed NLP models on \dataname{} to examine agenda setting, framing, and priming, several strategies underlying information manipulation, and reveal variation across media outlet control, social media platform, and time.
Our examination of these media effects and extensive discussion of current approaches' limitations encourage further development of NLP models for understanding information manipulation in emerging crises, as well as other real-world and interdisciplinary tasks.
\end{abstract}

\section{Introduction}
\label{sec:intro}

On February 24, 2022, Russia began an open military invasion of Ukraine. At the time of writing, this ongoing conflict has killed thousands of people and displaced millions.\footnote{
\href{https://www.unhcr.org/en-us/news/press/2022/3/622f7d1f4/private-sector-donates-us200-million-unhcrs-ukraine-emergency-response.html}{UNHCR}, 
\href{https://www.cbsnews.com/news/ukraine-russia-death-toll-invasion/}{CNN}}
The conflict has also manifested in ongoing \textit{information warfare}, as Russian, Ukrainian, and ally forces attempt to shape online narratives of the war.\footnote{\href{https://www.theatlantic.com/technology/archive/2022/03/russia-ukraine-war-propaganda/626975/}{Atlantic} 
}
Even before these events, Russian-backed entities have aimed to influence opinions outside \citep{arif2018acting,starbird2019disinformation}  and inside Russia \citep{field2018framing,golovchenko2018state,rozenas2019autocrats}.
More broadly, researchers consider the manipulation of public opinion over social media as ``a critical threat to democracy'' and have identified computational propaganda in over 80 countries \citep{bradshaw2021industrialized}.

\begin{figure}
    \centering
    \includegraphics[width=\columnwidth]{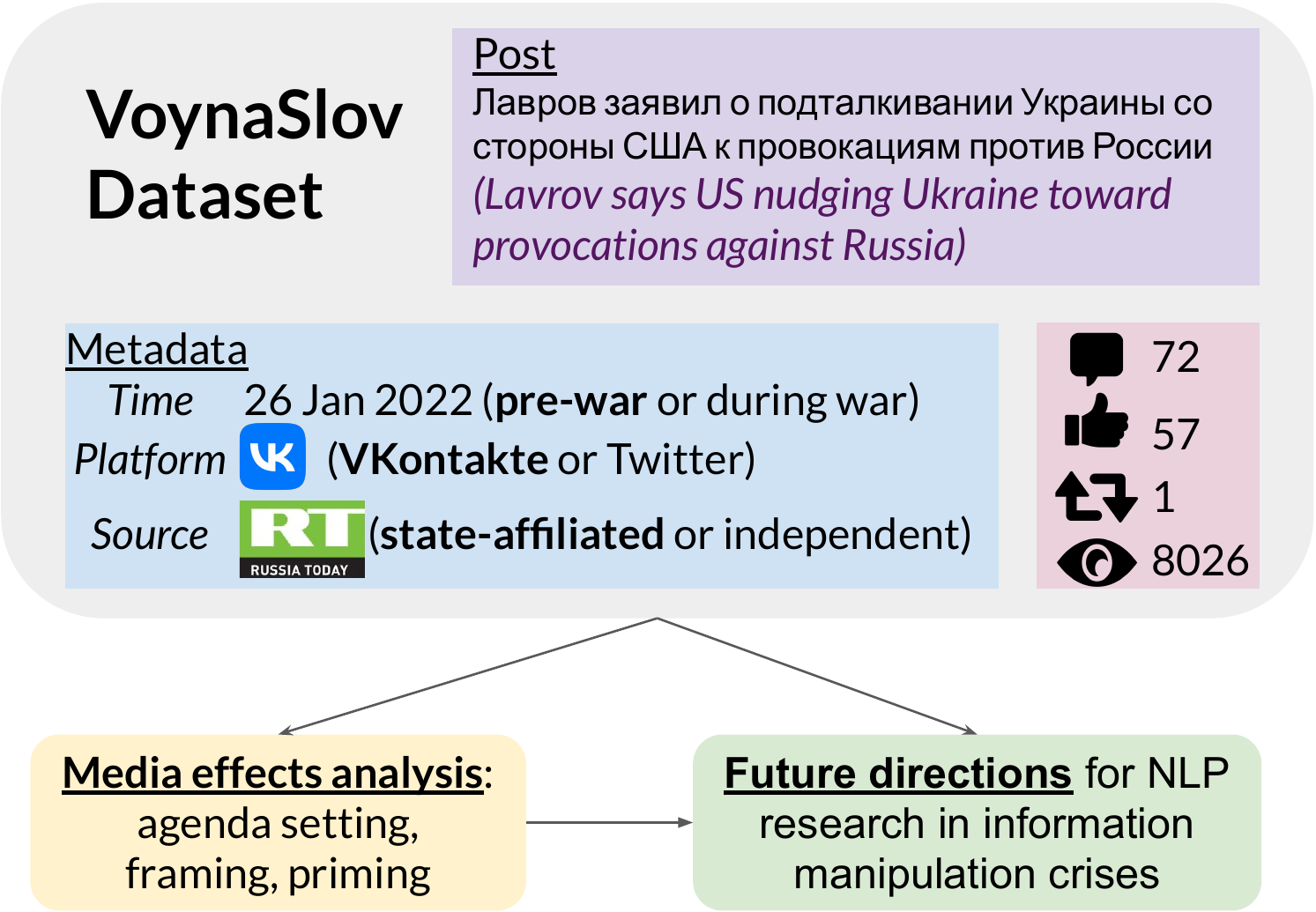}
    \caption{We create \dataname{}, a dataset of Russian news organizations' social media posts and public responses. Above is an example VK post in \dataname{} from the state-affiliated outlet \textit{Russia Today}, along with metadata and engagement metrics. We use \dataname{} to analyze media effects of agenda setting, framing, and priming. Both the dataset and our analyses contribute to our discussion of future directions for NLP research.}
    \label{fig:intro-fig}
\end{figure}

Because these campaigns often rely on text-based news and social media content, NLP can be a valuable tool in combating them.
In this work, we examine the usability of NLP approaches for combating information manipulation campaigns through the release of an in-progress data set, \dataname{}, focused on the 2022 Russia-Ukraine war (\Sref{sec:data}).
This dataset itself addresses one challenge for research in this space: the need for real-world data.
In contrast to contemporaneous Twitter data sets \citep{haq2022twitter,chen2022tweets}, our corpus is explicitly designed to capture Russian-government-backed information manipulation; we collect posts by state-affiliated and independent Russian media outlets and reactions to them on Twitter, which is more dominant in Europe and the U.S., and VKontakte (VK), one of the most widely-used social media platforms in Russia \cite{makhortykh2017social}.\footnote{\href{https://www.linkfluence.com/blog/russian-social-media-landscape}{https://www.linkfluence.com/blog/russian-social-media-landscape}\newline
The data, available at \href{https://github.com/chan0park/VoynaSlov}{https://github.com/chan0park/VoynaSlov}, currently contains $>$38M posts and will continue to be updated.}
Unlike approaches that crowd-source data to study a specific task \citep{thorne-etal-2018-fever,perez-rosas-etal-2018-automatic}, we derive research tasks directly from real-world data.

The dominant NLP research paradigm in information campaigns has focused on automated fact-checking or propaganda and fake news detection \citep{thorne-etal-2018-fever,oshikawa-etal-2020-survey,ijcai2020-0672,Zhou2020,guo-etal-2022-survey}.
However, this work typically involves supervised approaches and pre-annotated data, which are not available in emerging situations, and only captures one extreme form of media manipulation.
Instead, we draw on a common paradigm of information manipulation from communications research and examine signs of \textit{agenda setting}, \textit{framing}, and \textit{priming} in \dataname (\Sref{sec:analysis}).
For each media effect, we investigate the utility of the most common and recently developed NLP approaches.
Our analysis first reveals evidence of manipulation tactics in our data, showing that \dataname{} presents an avenue for studying them, and second, exposes open challenges in current NLP approaches towards uncovering, analyzing, and mitigating information manipulation campaigns.

We conclude by highlighting broader limitations of extant NLP approaches, discuss why model performance advancements have not yet translated to deployable technology in crises, and propose directions for future work to close this gap  (\Sref{sec:open}). 
Our contributions, visualized in Figure \ref{fig:intro-fig}, are a new data set of Russian media activity, which we use to analyze media effects, and an in-depth discussion of challenges and opportunities in NLP research on information manipulation campaigns. We hope to facilitate research on information warfare and ultimately enable reduction and prevention of disinformation and opinion manipulation.

\section{\dataname}
\label{sec:data}

\dataname contains posts from Russian news outlets on VK and Twitter, which primarily feature breaking news or summaries of original articles. 
Here, we describe data collection and statistics.

\paragraph{List of News Outlets}
\label{subsec:news-outlets}
We identified Russian media outlets and their Twitter and VK handles starting from a seed list.\footnote{\href{https://en.wikipedia.org/wiki/Mass\_media\_in\_Russia}{``Mass media in Russia'', Wikipedia}} We then selected other media accounts followed by the seed outlets on Twitter, repeating until convergence. Twitter identifies state-affiliated Russian media accounts with a badge\footnote{\href{https://help.twitter.com/en/rules-and-policies/state-affiliated}{About government and state-affiliated...labels on Twitter}}, which we use to label outlets as \emph{state-affiliated} or \emph{independent}.
The resulting list was manually verified by a fluent Russian speaker and includes 23 state-affiliated and 20 independent outlets (\Aref{appendix:media-handles}). However, we note that independent outlets may not be truly independent from state influence, particularly due to restrictions on free speech since the invasion.\footnote{\href{https://www.nytimes.com/2022/03/04/world/europe/russia-censorship-media-crackdown.html}{Russia Takes Censorship to New Extremes, Stifling War Coverage (NYT)}}
We collect data as early as January 2021, over a year before the war, as many believe the invasion was planned far in advance and the media may have preemptively planted narratives.\footnote{e.g., \href{https://en.wikipedia.org/wiki/Prelude\_to\_the\_2022\_Russian\_invasion\_of\_Ukraine}{``Prelude to the 2022 Russian invasion of Ukraine''}}

\begin{table}[t]
\centering
\resizebox{0.98\columnwidth}{!}{
\begin{tabular}{l|cc|cc}
                          & \multicolumn{2}{c|}{\textbf{Media Posts}} &  \multicolumn{2}{c}{\textbf{Public Reac.}} \\
                          & SA              & Ind             & SA               & Ind              \\\hline
\textbf{VK} (Pre-war)        &     333K          &      143K         &      11M            &     3M          \\
\textbf{VK} (Wartime)      &      94K           &     27K          &      6M            &      430K         \\\hline
\textbf{Twitter} (Pre-war)   &        41K        &        33K       &      -            &     -           \\
\textbf{Twitter} (Wartime) &     109K            &     36K          & \multicolumn{2}{c}{{17M}}                    \\\hline
\end{tabular}
}
\caption{Number of posts/comments/tweets by state-affiliated (SA) and independent (Ind) media in \dataname.
}
\label{tab:total-stats}
\end{table}

\paragraph{\datanamevk}
\label{subsec:data-collection-vk}

We collect VK posts from identified media accounts with the VK Open API.\footnote{\href{https://vk.com/dev/openapi}{https://vk.com/dev/openapi}
All VK media account pages are publicly available.
Our data release provides only posts/comment IDs to abide by VK's terms and conditions. Full data can be restored using VK Open API as long as it remains available at the time of collection.}
\Tref{tab:total-stats} provides a detailed breakdown of the 21M+ posts collected.
For each post, we collect the number of views, likes, the presence of images, videos, and links, and comments to capture \textit{public reaction}.

\paragraph{\datanametwitter}
\label{subsec:data-collection-twitter}

We similarly collect tweets and metadata such as like and retweet count from Russian media accounts.
We capture public reaction with the Twitter search API and iteratively craft search terms. Starting from a small seed list, we collect an initial set of tweets. We augment our seed list with frequent terms from this initial set judged to be relevant to the war. After several rounds, our final list contains 264 terms and hashtags (\Aref{appendix:twitter-search-terms}). Since the Twitter API only supports search with a 7-day limit, we collect 17M public tweets from 24 Feb - 31 May 2022. 

\paragraph{Data Statistics}
Table \ref{tab:total-stats} presents basic data statistics. We provide additional metrics in \Aref{appendix:data-stats} and summarize here.
Due to Twitter's 280 character constraint, VK posts tend to be longer than tweets, and independent media posts are significantly longer than state-affiliated posts on both platforms.
Compared to independent outlets, state-affiliated outlets include much more multimedia, which can be powerful framing devices \citep{powell2015clearer}.
Most state-affiliated and independent tweets include external links, which enhance users' perceptions of trustworthiness and credibility \citep{morris2012tweeting,wang2013trust}.
However, there is a stark difference on VK, where 76.3\% of independent media posts include external links compared to just 26.5\% of state-affiliated posts.

\dataname{} suggests that state-affiliated media dominates VK, but independent media dominates Twitter. 
On average, state-affiliated VK accounts have 26K posts, over twice as much as independent media. This pattern is reversed on Twitter, where independent accounts are slightly more active. 
VK's publicly-available data includes view counts, presenting a unique opportunity to study exposure \citep{tewksbury2001accidentally}. Not only are state-affiliated outlets more active on VK, but their content reaches a larger audience (18K vs 10K views per post).

Popularity cues, e.g., likes, comments, and retweets, can serve as indicators of the success of the media's opinion manipulation strategies (see \Sref{sec:priming}). 
Independent media posts receive more engagement on Twitter, but state-affiliated posts receive more engagement than independent posts on VK. However, independent posts on VK still have engagement rates if we account for their smaller audiences.
As discussed in \Sref{sec:intro}, VK is more widely used in Russia and enables analyzing internal Russian information manipulation campaigns as well as reactions of people likely to be in Russia.
Our data enables comparisons between VK and Twitter and could reveal differences in strategies used by state-affiliated Russian media when targeting domestic and international audiences.

\section{Facilitation of NLP Research on Media Opinion Manipulation}
\label{sec:analysis}
We demonstrate how \dataname can facilitate research on media opinion manipulation by focusing on three media effects: agenda-setting (\Sref{sec:agenda_setting}), framing (\Sref{sec:framing}), and priming (\Sref{sec:priming}), though we note disputes over the distinctness and convergence of these concepts \citep{price1997news,Dietram2000,ghanem2001convergence}.
For each media effect, we first provide background and review existing NLP approaches. We then apply current state-of-the-art models from the most dominant NLP paradigms on \dataname, presenting evidence that this data can support examinations of these effects. We conclude each subsection with open challenges exposed by our analyses.

\subsection{Agenda Setting}
\label{sec:agenda_setting}

\paragraph{Background}

\textit{Agenda setting}, first introduced by \citet{McCombs1972}, suggests that the importance attributed to issues by audiences is strongly correlated with the emphasis that mass media place on them \citep{scheufele2007framing}.
An actor seeking to manipulate public opinion can influence  how important an audience considers specific issues by reducing or increasing their representation in the media.
As news topics are event-driven and agenda setting strategies are unknown in a new corpus, NLP approaches to uncovering them use statistical and unsupervised methods, including word statistics or Bayesian models and evaluating against external indicators of events \citep{tsur-etal-2015-frame,field-etal-2018-framing}.

\paragraph{Results in our Data}

\begin{figure}
    \centering
    \includegraphics[width=0.5\textwidth]{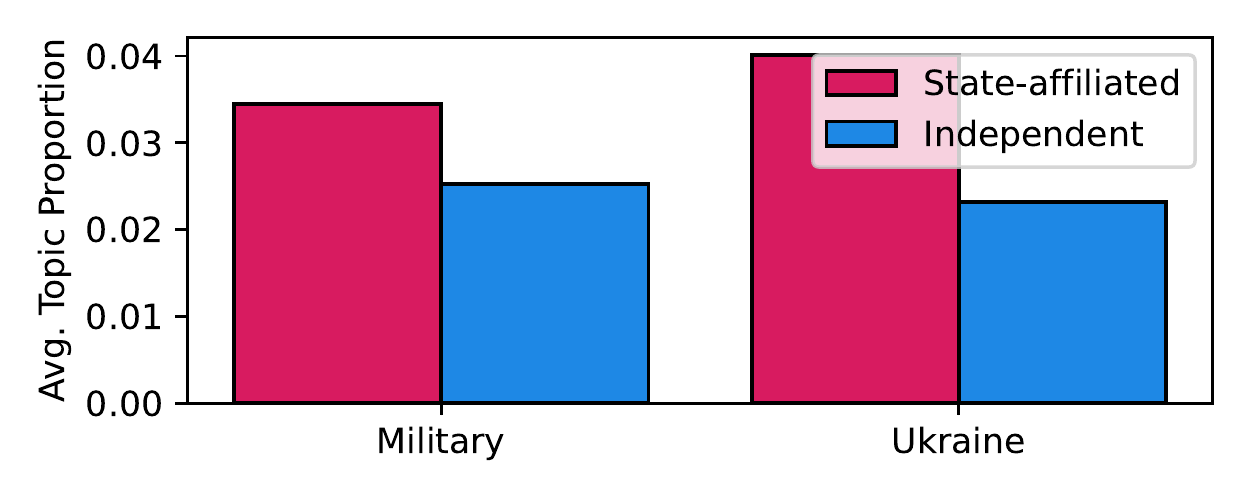}
    \caption{War-related topic proportions for state-affiliated and independent media outlets, as learned by a 30-topic CTM. We display the two topics we identified as most war-related: ``Military'' (top words: \russian{округа}:area, \russian{авиации}:aviation, \russian{военного}:military, \russian{военно}:military, \russian{флота}:navy, \russian{учения}:military exercise, \russian{су}:Sukhoi~Su, \russian{военнослужащие}:military personnel, \russian{противника}:enemy, \russian{сил}:forces), and ``Ukraine'' (top words: \russian{днр}:DPR, \russian{лнр}:LPR, \russian{мариуполя}:Mariupol, \russian{мирных}:peaceful, \russian{жителеи}:residents, \russian{украинские}:Ukrainian, \russian{народнои}:folk, \russian{новости}:news, \russian{донбасса}:Donbass, \russian{мариуполе}:Mariupol). We report all topics in \Fref{fig:neural_all}.}
    \label{fig:neural_select}
\end{figure}

\dataname facilitates examination of agenda setting by including posts from various outlets and labels of outlets as state-affiliated or independent.
Because we need unsupervised analyses of \textit{what} topics are covered, we identify topic modeling and word frequencies as the most imminently usable NLP methods.

We employ two different topic models: a structured topic model  \citep[STM]{Roberts2016,Roberts2019} and a contextualized neural topic model \citep[CTM]{bianchi-etal-2021-pre,bianchi-etal-2021-cross}. The STM is a popular LDA-style probabilistic model that improves upon prior approaches by allowing users to incorporate arbitrary metadata.
The CTM is based on a  variational autoencoder \citep{srivastava2016autoencoding} and appends pre-trained sentence embeddings \citep{reimers-gurevych-2019-sentence} to bag-of-words document representations, reducing the bag-of-words assumptions made by traditional models. We train both models over VK posts from January 1, 2021 to May 15, 2022.
For the STM, we include affiliation (state or independent) and time (days) as topic prevalence covariates. \Aref{appendix:topic_model_params} provides details.

\Fref{fig:neural_select} shows selected topic proportions estimated by the CTM, averaged over posts from state-affiliated and independent outlets. \Aref{appendix:topic_model_params} reports full results for both models.
Both models show differences in topic distributions in state-affiliated and independent outlets, suggesting this data offers opportunities for examining how coverage differs by outlet affiliation.

\begin{figure}
    \centering
    \includegraphics[width=0.5\textwidth]{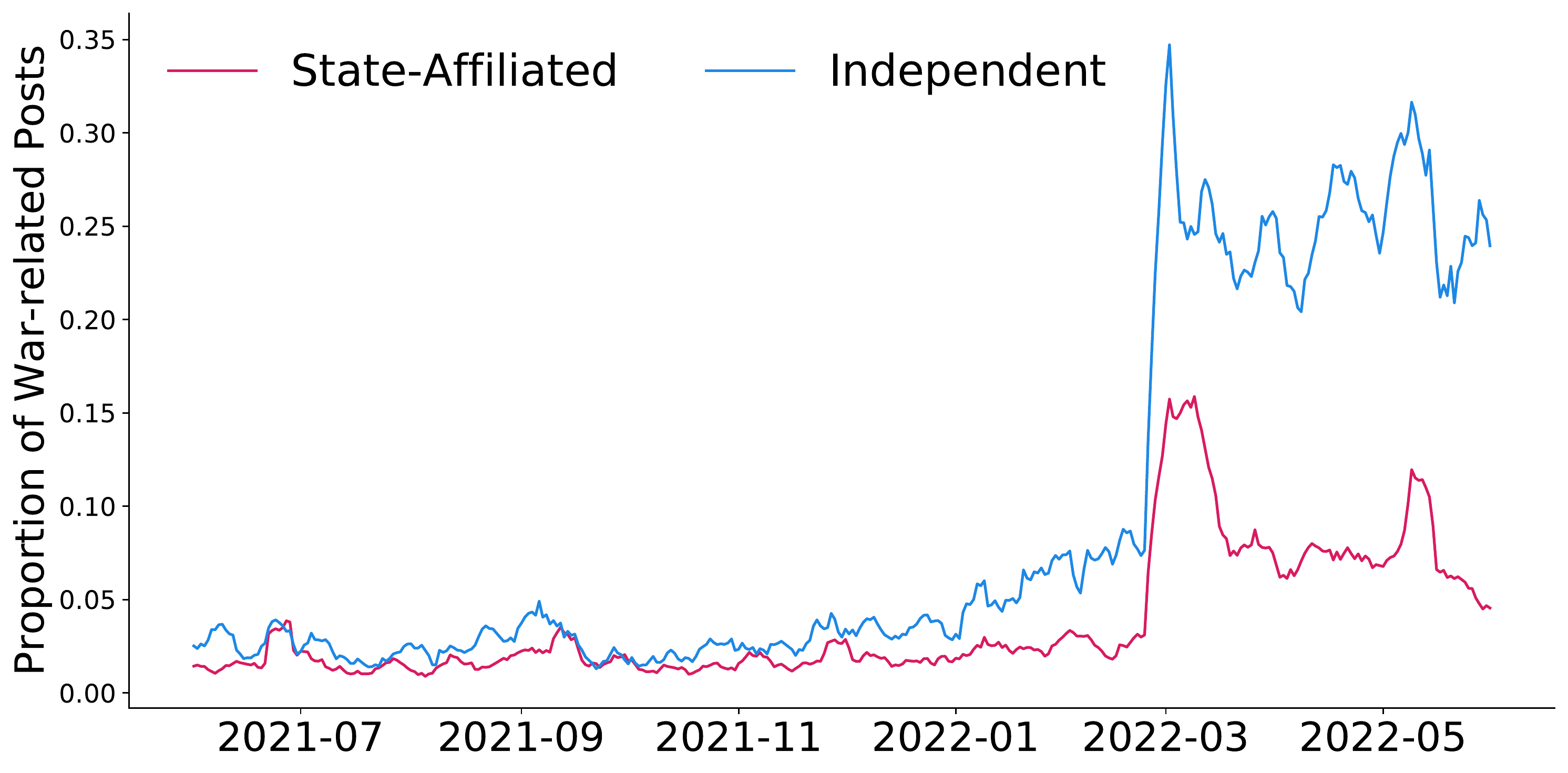}
    \caption{Proportion of posts that mention war-related terms (e.g. ``war'', ``operation'') in state-affiliated and independent outlets over time. Independent VK posts use these terms more frequently, especially following the invasion.}
    \label{fig:word_freq_war}
\end{figure}

Many speculate the extent of state-affiliated media's war coverage and suggest that people in Russia have little knowledge of the invasion.\footnote{Examples: \href{https://time.com/6179221/putin-information-war-column/}{Time},
\href{https://www.cnn.com/2022/04/03/media/russia-media-ukraine-cmd-intl/index.html}{CNN}}
Omitting coverage constitutes agenda setting. The most war-related topics are CTM topics 5 and 14 (\Fref{fig:neural_select}) and STM topic 19 (\Fref{fig:stm_all}); the prevalence of these topics sharply increases in both types of news outlets in late February (\Aref{appendix:topic_model_params}). However, these three topics have higher prevalence estimates in state-affiliated than independent media. In contrast, \Fref{fig:word_freq_war} uses word statistics to more directly examines how often each media type mentions the war. A much higher proportion of independent posts mention war-related terms (e.g., ``war'', ``operation''), especially since the invasion, which is consistent with our observations from manually reading randomly sampled posts.

\paragraph{Open Questions}
Based on these results, we highlight 3 main limitations in current approaches (topic modeling and word statistics) for uncovering agenda-setting strategies, \textit{uninpretability}, \textit{instability}, and \textit{over-simplification}.
The difficulty of interpreting topic modeling and related approaches has long been acknowledged (e.g., \citep{Chang2009Tea}) and remains an open challenge despite the continued popularity of these methods. In our data, not all topics are coherent, and even in coherent topics, comprehensiveness is difficult to determine. For example, the most Ukraine-related CTM topic (Topic 5) references the two largely-unrecognized breakaway states in eastern Ukraine: \textit{DNR} and \textit{LNR}, suggesting that this topic captures explicitly pro-Russia coverage of events, which we expect to be more common in state-affiliated outlets. There is no straightforward mechanism to prevent or easily recognize the one-sidedness of topics.

Relatedly, results vary even under similar models. Topic modeling is sensitive to pre-processing decisions \citep{denny_spirling_2018}, and word frequencies depend entirely on the choice of words. While word statistics and manual analysis show evidence that independent outlets discuss the war more frequently than state-affiliated outlets, topic models suggest the opposite. 
Instability makes results difficult to trust, and more research is needed to improve consistency and reliability.

Finally, simplifying assumptions likely limit conclusions. Word-level metrics fail to account for context, and most topic models, including STMs, make bag-of-words and independence assumptions. While the CTM relaxes bag-of-words assumptions with sentence embeddings, a disadvantage compared with the STM is that it does not parameterize topics with metadata. Combining contextualized embeddings with flexible neural architectures could provide avenues for relaxing assumptions \citep{card-etal-2018-neural,Zhao2021}.
However, using embeddings pretrained on external data risks introducing false findings derived from the external data, rather than from the target analysis corpus \citep{field-tsvetkov-2019-entity,shwartz-etal-2020-grounded}.

Identifying agenda setting requires examining topics in unseen corpora. Alternative methods do exist, such as embedding clustering \citep{sia-etal-2020-tired}, and if researchers have specific hypotheses, hand-coding articles or constructing lexicons may be possible. Nevertheless, unsupervised word-level metrics remain go-to approaches for topic analysis, and interpretability, stability, and reducing simplifying assumptions remain open challenges.

\subsection{Framing}
\label{sec:framing}

\paragraph{Background}

In media studies, whereas agenda setting refers to \textit{what} topics are discussed, framing is based on the assumption that \textit{how} those topics are discussed can influence the way audiences understand them \citep{scheufele2007framing}.
Framing has origins in sociology \citep{goffman1974frame} and psychology \citep{tversky1981framing} as well as communication \citep{entman1993framing}. Psychologists tend to focus on \textit{equivalence} frames: different presentations of logically-identical information \citep{scheufele2012state}, such as using the phrases ``90\% employment'' vs. ``10\% unemployment'' \citep{chong2007framing,tewksbury2019news}. In contrast, \textit{emphasis} frames present ``qualitatively different yet potentially relevant considerations'' \citep[~p.114]{chong2007framing}, such as focusing on free speech vs. public safety in new coverage of a protest.
Frames can also be \textit{issue-specific}, which facilitates highly detailed analyses, or \textit{generic}, which facilitates replicability and generalizablility \citep{devreese2005news}.

Much NLP research has focused on detecting \textit{generic} frames, using the Media Frames Corpus (MFC) \citep{card-etal-2015-media,johnson2017ideological,field2018framing,khanehzar-etal-2019-modeling}, moral foundations \citep{mokhberian2020moral,roy2021identifying}, or episodic and thematic frames \citep{mendelsohn2021modeling}. Due to high expert annotation costs, there has been considerably less attention to \textit{issue-specific} frames, but several recent works showcase how they can enrich our understanding of discourses \citep{morstatter2018identifying,liu2019detecting,mendelsohn2021modeling}. 
NLP research also mirrors the debate over whether frames should be identified inductively as they emerge from the data under study, or a-priori based on existing theories in a deductive manner \citep{devreese2012news}: some models are (sometimes weakly) supervised \citep{morstatter2018identifying,khanehzar-etal-2019-modeling,roy-goldwasser-2020-weakly}, while others are unsupervised \citep{kwak2021frameaxis,nicholls2021computational,yu2021frame}.

\dataname offers unique opportunities to study \textit{frame-building}, how social forces (e.g., organizational pressures or journalists' ideologies) influence what frames are cued by media coverage \citep{devreese2005news}, along several dimensions, notably ownership (state-affiliated or independent), platform (Twitter or VK), and time.
We first compare two \textit{issue-specific equivalence} frames: use of words denoting ``war'' vs.~the euphemism ``military operation''. Then, we develop a state-of-the-art model to analyze \textit{generic emphasis} frames from the MFC \cite{card-etal-2015-media}.

\paragraph{Results in our Data}

\begin{figure}
    \centering
    \includegraphics[width=\columnwidth]{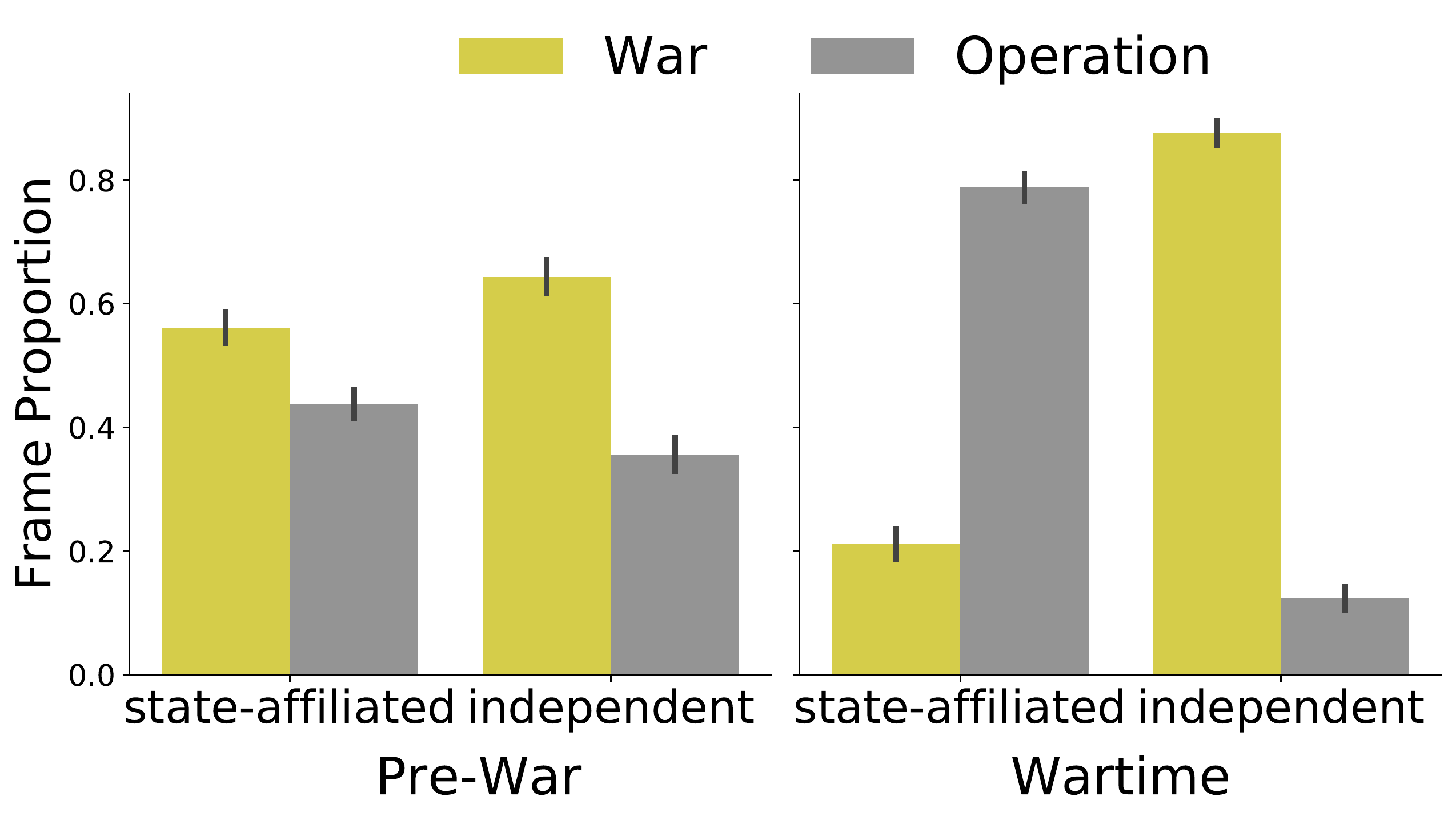}
    \caption{Proportion of media posts containing the ``war'' vs. ``operation'' equivalence frames, normalized by the number of posts using either. Since the war started, independent outlets have used the ``war'' frame much more frequently, while state-affiliated outlets exhibit a strong preference for the ``operation'' frame.}
    \label{fig:war_op}
\end{figure}

Whereas \Fref{fig:word_freq_war} depicts how often outlets mention the war at all, Figure \ref{fig:war_op} focuses on what terminology they use by examining ``war'' vs. ``operation'' \textit{issue-specific equivalence} frames. Since the onset of the war, independent outlets have exhibited a strong preference for ``war'' while state-affiliated ones more often use the ``operation'' euphemism.
These findings are consistent with other accounts describing that the Russian government downplays the severity and aggression of the invasion and eventually even banned media from using the terms ``war" and ``invasion".\footnote{\href{https://www.cbsnews.com/news/russia-ukraine-war-what-do-russians-see-and-hear-at-home/}{CNN}}

Next, we train sentence-level classifiers to detect 15 \textit{generic emphasis} frames using the annotated MFC, standard data for frame analysis in NLP \cite{card-etal-2015-media}.
We construct classifiers based on large pre-trained language models shown to be the state-of-the-art \citep{kwak2020systematic, akyurek-etal-2020-multi}. 
Unlike prior work, which evaluates in-domain, we more realistically simulate both in-domain and zero-shot scenarios with the MFC, and also evaluate with \dataname. As the MFC is organized by policy issue, we simulate zero-shot classification by leaving one issue as a test set (e.g. immigration) and using remaining data for training and development (e.g., same-sex marriage and tobacco). To evaluate over \dataname, a native Russian speaker annotated randomly sampled sentences from \datanamevk.\footnote{See \Aref{appendix:model-training} for model and \ref{appendix:mfc-annotation} for annotation details.}

Unsurprisingly, performance significantly drops in the zero-shot setting within the MFC corpus, an even more so in \dataname, which features content in a different language\footnote{When we apply the MFC classifier over VoynaSlov for both evaluation and analysis, we translate original Russian texts to English as MFC only contains English data.}, cultural context, style, and format than the MFC news articles (\Tref{tab:res_framing}). Nevertheless, we analyze frame-building with the predicted MFC frames.\footnote{See \Aref{appendix:mfc-label-generation} for more information.} Following \citet{mendelsohn2021modeling}, we fit separate mixed-effects logistic regression models with each frame's presence as a binary dependent variable. Fixed effects include ownership (state-affiliated vs. independent) and platform (Twitter vs. VK), and we control for specific media outlet and date as random effects.

\begin{table}[]
\centering
\resizebox{0.85\columnwidth}{!}{
\begin{tabular}{llll}
                            & \textbf{Data}     & \textbf{Model} &  \textbf{F1} \\\hline
In-domain                   & MFC               & XLM-R$_{L}$    &  67.5        \\\hline
\multirow{4}{*}{Zero-shot}  & Immigration       & XLM-R$_{L}$    &  52.7        \\
                            & Same-sex          & XLM-R$_{L}$    &  50.4        \\
                            & Tobacco           & XLM-R$_{L}$    &  51.0        \\\cline{2-4}
                            & VoynaSlov         & XLM-R$_{L}$    &  33.5       \\\hline 
\end{tabular}
}
\caption{Macro-F1 results of trained MFC classifiers in in-domain, zero-shot, and \dataname setups.}
\label{tab:res_framing}
\end{table}

\begin{figure}
    \centering
    \includegraphics[width=\columnwidth]{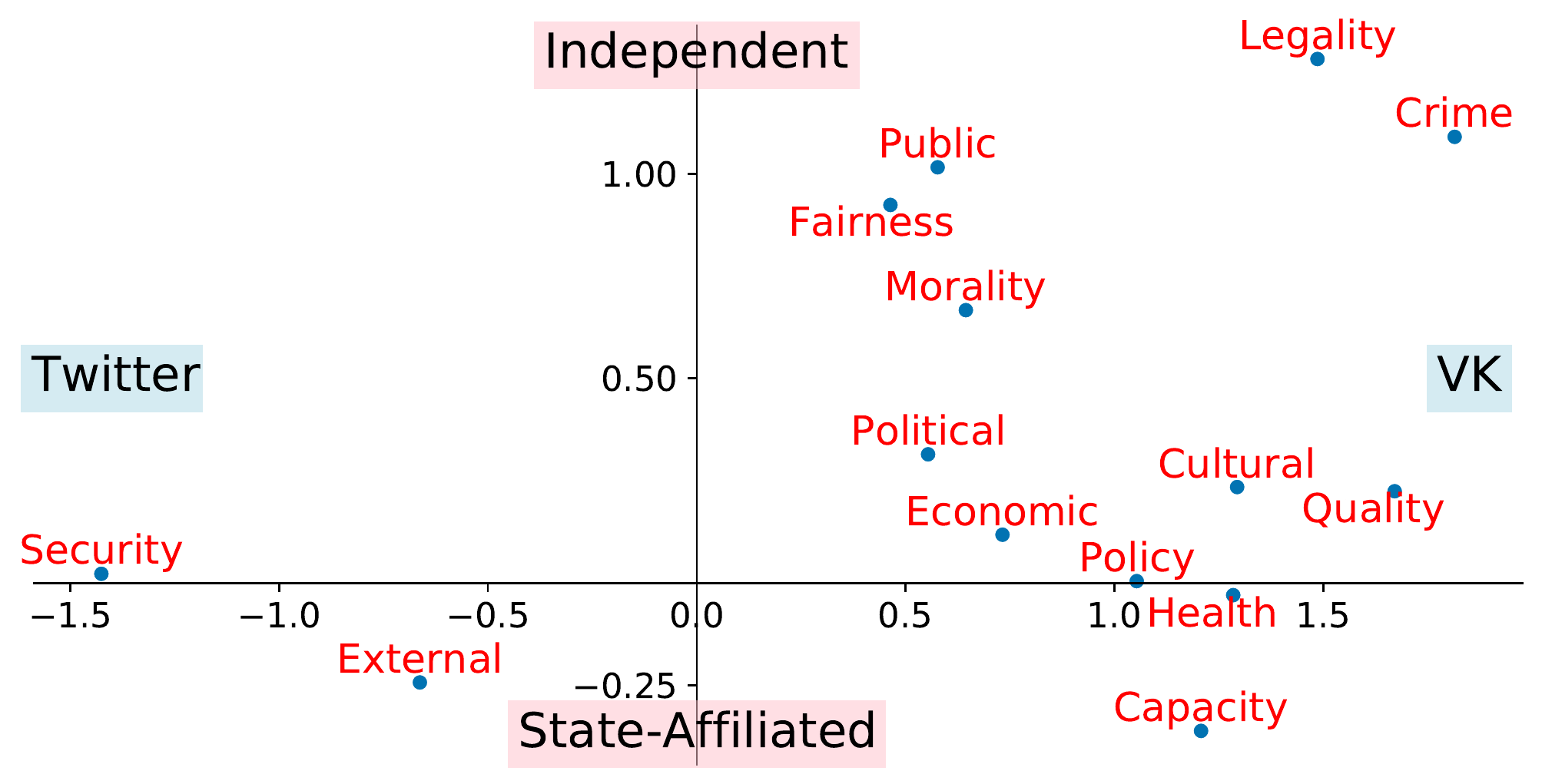}
    \caption{Association between framing and media ownership (y-axis; independent vs. state-affiliated) and social media platform (x-axis; Twitter vs. VK). The plotted values represent $\beta$ coefficients when using ownership and platform as features in a logistic regression model to predict the presence of each frame. For example, the ``crime'' frame (upper right) is most strongly associated with independent media and VK. All values with magnitude $>0.5$ are significant ($p < 0.01$).}
    \label{fig:mfc_build}
\end{figure}

Cued MFC frames vary across  platform and media ownership (\Fref{fig:mfc_build}). Compared to state-affiliated outlets, independent outlets are more likely to use \textit{Legality} and \textit{Crime \& Punishment}, which possibly indicates questioning legal precedent of the invasion or criminal activity of the military. Frames that capture human rights (\textit{Fairness \& Equality}, \textit{Morality}) and citizens' views (\textit{Public Sentiment}) are also significantly associated with independent media. Regarding platform effects, most frames are associated with VK, which may reflect lack of platform character limits: VK enables more in-depth posts that explicitly cue MFC frames. However, \textit{External Regulation \& Reputation} and \textit{Security \& Defense} are used significantly more on Twitter. Since both frames face ``outward'' by focusing on Russia's relationships with other countries, this result supports our speculation that Russian media use Twitter to reach people outside of Russia, and VK to reach people within Russia. However, it is difficult to draw more specific conclusions because \Fref{fig:mfc_build} does not reveal what each frame means in the context of \dataname{}; indeed, the MFC typology may obscure more meaningful framing patterns \citep{mendelsohn2021modeling}.

\paragraph{Open Questions}
We focus on two open questions exposed by our data and analysis: \textit{Unclear typology} and \textit{Domain-specificity}.
Even in other disciplines, there is no consensus on whether frames should be specific or generic, equivalence or emphasis, and inductive or deductive, and not only what typology is appropriate for specific research questions and corpora, but also what best reflects the psychology processes by which people are actually influenced by framing \citep{devreese2012news}.
Generic emphasis frames have become the dominant typology in NLP, likely because this approach aligns with standard NLP paradigms of classification and reusable data.
However, the difficulty of interpreting \Fref{fig:mfc_build} suggests the MFC frames may not be the most relevant in \dataname, despite this being the easiest typology to imminently operationalize. 
While less-supervised approaches exist, they often use word statistics or hierarchical topic models, which lead to the same interpretability, stability, and simplification concerns as in \Sref{sec:agenda_setting} \citep{Nguyen2013,Roberts2016,demszky-etal-2019-analyzing,bhatia-etal-2021-openframing}.

Even with an established typology, framing is highly context-dependent, as it is a ``bridging concept between cognition and culture'' \citep[~p.61]{vangorp2007constructionist}.
The need to capture subtle and nuanced content is an ongoing challenge in NLP research: models often overfit to shallow lexical features and generalize poorly to new domains \cite{daume2006domain}, which \Tref{tab:res_framing} exemplifies.
While domain-adaption is a large field in NLP, it is unclear how well these approaches work in detecting nuanced concepts, and model complexity may reduce deployability. Although surfacing framing strategies remains challenging, particularly in an emerging crisis outside of the U.S. political context, Figures \ref{fig:war_op}-\ref{fig:mfc_build} show signs of \textit{frame-building} in our corpus, suggesting that \dataname{} offers avenues for future framing research.

\subsection{Priming}
\label{sec:priming}

\paragraph{Background}
Priming typically refers to the effects of framing and agenda setting \citep{entman2007framing}.
Some researchers use the term to specifically refer to  ``changes in the standards that people use to make political evaluations'' \citep{iyengar1987news}, which is associated closely with agenda setting \citep{scheufele2007framing,Moy2016}. For example, news coverage of particular issues encourages audiences to base judgements of leaders and governments on these issues at the exclusion of others, including during elections \cite{scheufele2007framing}.
The effects of framing, or audiences' adoption of frames presented in news as ways to understand issues, can then be termed \textit{frame setting} \citep{Moy2016}.
We take a broad definition of the term \textit{priming} and consider both agenda setting and framing effects in this section.

Little work in NLP has focused on priming. Some aspects of how readers respond to news coverage fall outside the scope of text analysis \citep{zubiaga2016analysing} and may be better examined through surveys, historical polls, or election results \citep{price1997switching,valkenburg1999effects, zhou2007parsing}. User reactions on social media, including likes, shares, and comments, can also offer some insight into how framing and agenda setting strategies are received. 
In this section, we show that the inclusion of reactions in \dataname offers an avenue for studying priming, but that this line of research raises technical and ethical challenges.

\paragraph{Results in our Data}

We investigate the effects of MFC frames on user engagement with mixed-effects linear regression models. Independent variables include the presence of each frame, ownership (state-affiliated or independent), and if a post has an image, video, or link (each coded as binary factors). Random effects include specific outlet and date. We consider four outcomes: numbers of views, likes, and reposts (all log-scaled), and engagement rate, defined as the sum of like, repost, and comment counts normalized by the view count. 

\Tref{tab:frame-setting} shows the variables most strongly associated with each engagement metric. 
Including multimedia, especially videos, is strongly predictive of user engagement. Civilian-focused frames (e.g. \textit{Public Sentiment}, \textit{Morality}), are linked to higher engagement in all four measurements. However, as in \Sref{sec:framing}, these correlations are sometimes difficult to interpret. For example, \textit{Policy} is most strongly associated with more reposts, but we are unable to decipher what this frame captures and what the implications may be for media manipulation campaigns.

\begin{table}[]
\resizebox{\columnwidth}{!}{%
\begin{tabular}{@{}llll@{}}
\toprule
Views                   & Likes                   & Reposts                & Engagement               \\ \midrule
Has Video (.24)        & Has Video (.56)        & Policy (.40)          & Public Sent. (.31) \\
Has Image (.20)        & Public Sent. (.35) & Has Video (.37)       & Has Video (.23)        \\
Public Sent. (.13) & Morality (.27)         & Morality (.28)        & Morality (.17)         \\
Crime (.12)            & Security (.21)         & Qual. of Life (.24) & Has Link (.16)         \\
Fairness (.11)         & Has Image (.18)        & Capacity (.22)        & Security (.08)         \\
\end{tabular}%
}
\caption{Frame-setting results on engagement metrics. The numbers in parentheses indicate each feature's coefficient of trained regression models.}
\label{tab:frame-setting}
\end{table}

\begin{figure}
    \centering
    \includegraphics[width=\columnwidth]{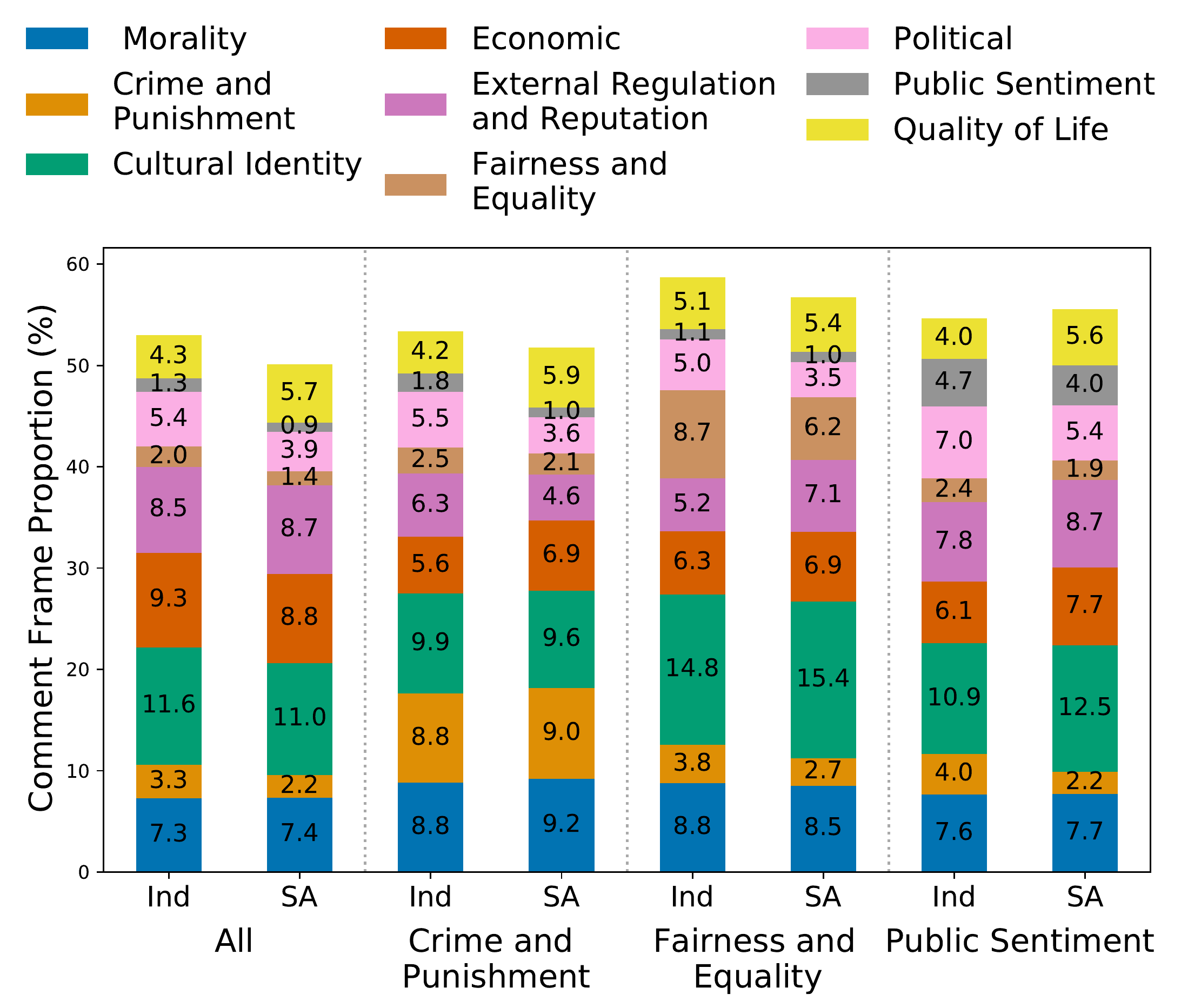}
    \caption{Comparison of frame proportion of comments of independent (Ind) and state-affiliated (SA) media, for each frame used by posts.}
    \label{fig:priming_frames}
\end{figure}

We further investigate how frames adopted by media posts affect readers' frame usage.
\Fref{fig:priming_frames} shows the average frame proportion of user comments, depending on the frames of original posts and media state-affiliation.
On average, users leave significantly fewer comments with \textit{Political}, \textit{Public Sentiment}, and \textit{Fairness \& Equality} frames on state-affiliated media, which might be related to Russian laws imposing strict censorship. Instead, comments on state-affiliated posts more often employ \textit{Economic} and \textit{Quality of Life} frames which might reflect what readers prioritize or feel comfortable discussing.

\paragraph{Open Questions}
We identify three primary open questions: \textit{Data validation}, \textit{Privacy}, and \textit{Technology Misuse}.
While we investigate priming through user reactions and comments, this approach is fundamentally limited. Priming relates to cognitive influence on individual users, which is impossible to capture through aggregate metrics. Additionally,
creating fake or disingenuous accounts to promote particular content is a known information manipulation strategy \citep{king_pan_roberts_2017,arif2018acting}, and studying unverified aggregated data could be capturing content by these accounts.  
Even if posts are made by genuine social media users, we cannot determine how reflective social media activity is of true cognitive states, individual attitudinal changes, or macro-level public opinion shifts. 

Deeper investigations of framing and agenda setting effects on individual users could better address this data validation issue, but raises concerns of \textit{privacy} and the potential for \textit{user-targeting}. 
Russian state surveillance of social media activity is well-established and there have been cases of civilians being arrested based on social media activity \citep{mejias2017disinformation,gabdulhakov2020trolling}. Furthermore, big data and computational social science research has been confronted with ethical challenges due to the lack of informed consent from participants and unawareness of how their data is being used by researchers \citep{fiesler2018participant,lazer2020computational}. Privacy-preserving modeling of social phenomena surfaced by language remains an open challenge in NLP. Recent social science research has linked social media activity with traditional surveys that abide by principles of informed consent and ethical research \citep{eady2019many}, which could be a viable path forward in NLP as well. 

Finally, research on priming has greater \textit{misuse} potential than framing or agenda setting. Framing and agenda setting analyses identify manipulation strategies already in use, and thus are unlikely to inform malicious actors on how to generate propaganda.
In contrast, priming research could directly inform malicious actors on which strategies are effective, though this has not deterred political psychology work on priming (e.g. \citealp{conway2017donald}).
The misuse potential poses a paradox:  uncovering effective framing and agenda setting strategies is more important than ineffective ones, but analyzing their effectiveness can lead to wider adoption. 
These challenges make studying priming from an NLP perspective difficult.
We suggest that \dataname facilitates research on understanding some aspects of the visibility and adoption of framing and agenda setting, but that more in-depth analyses may fall outside the scope of NLP research.

\section{Discussion}
\label{sec:open}

There are numerous opportunities for NLP research to have positive impacts in identifying and mitigating information manipulation campaigns. As a first step, we release \dataname, focused on Russian media activity on social media before and during the 2022 Russia-Ukraine war. Grounded in political communication research on media effects, we apply both traditional and state-of-the-art NLP techniques to analyze the language underpinning information manipulation. Indeed, we uncover variations in agenda setting and framing strategies across time (pre-war or wartime), social media platforms (VKontakte or Twitter), and media control (state-affiliated or independent). We encourage future work to continue to explore the plethora of social science theories and NLP techniques to analyze the data in \dataname. Furthermore, we hope that \dataname will aid future efforts in using NLP to address not only the 2022 Russia-Ukraine war, but emerging crisis situations more broadly. 

Through both our data collection and analysis, we learn that vast improvements in model performance on core NLP tasks have not yet translated into deployable technology capable of addressing information manipulation campaigns. Our work suggests several reasons for this discrepancy, and we foreground such limitations in order to set forth concrete directions for future work.
First, much prior work has focused on developing supervised models to detect fake news and propaganda in isolated media texts, including the establishment of shared tasks and standardized data sets \citep{thorne-etal-2018-fact,da-san-martino-etal-2020-semeval,shaar-etal-2021-findings}.
While there are settings where these approaches can be useful, they typically require carefully-labeled data that are unavailable in emerging settings, and we did not identify them as applicable in our analysis of \dataname.
Similarly, NLP advancements in model-pretraining \citep{Brown2020} are difficult to deploy over emerging data: most pre-training data is in English, and it is difficult to disentangle patterns in the target data from ones learned during pre-training \citep{field-tsvetkov-2019-entity,shwartz-etal-2020-grounded}.  
Pre-annotated data and pre-trained models can be immensely valuable for analyzing the past but have limited utility in understanding ongoing events. 

Second, even technological approaches that aim to target ongoing events, just as systems for aiding human fact-checkers \citep{Preslav2021} typically  consider only the most extreme and overt form of opinion manipulation: disinformation and fake news. However, our results build upon existing communications studies, demonstrating that media manipulation often constitutes more subtle strategies, such as selectively covering (or avoiding) issues (\Sref{sec:agenda_setting}) and changing minor word choices to influence audiences (\Sref{sec:framing}).
Little NLP work has examined \textit{agenda setting} and \textit{priming} at all. While substantial work has focused on \textit{framing} (\Sref{sec:framing}), it disproportionally focuses on U.S. politics, with few applications to non-English languages, other social contexts, or information warfare.

Third, NLP research on media manipulation has primarily examined isolated news texts without additional context, neglecting the larger hybrid media ecosystem comprised of intricate interactions between journalists, media organizations, political actors, social media platforms, and civilians \citep{chadwick2017hybrid}.
\dataname{} attempts to facilitate research in this area with content from a specific context and includes multiple outlets and platforms.
Nevertheless, it remains challenging to truly comprehend the media's motivations for how they present the news and their desired effects on public opinion, which then enables specific and nuanced analyses of manipulation strategies.   

Although we emphasize limitations in existing NLP approaches, we conclude by asserting that NLP has a unique opportunity to uncover information manipulation campaigns and contribute to social science research. Media effects have been rigorously studied within social science, but common approaches, including focused analyses of small sets of articles and human experiments with constructed stimuli in highly-contrived settings,  are insufficient for assessing the scale and societal impact of media manipulation. Computational methods can be representative of the full media environment and capture more realistic audience responses to news content shared on social media.

This work aims to shift the paradigm for research on automated opinion manipulation to encompass broader tactics, have grounding in social science theory, and incorporate emerging context. 
We believe these expansions will enable NLP to have positive impact \textit{during}, rather than \textit{after}, ongoing crisis situations.
We hope that our release of \dataname, our analysis of media effects, and our discussion of open NLP challenges facilitate the detection and ultimately the prevention of information manipulation.

\section{Limitations}

Our work includes the release of a new data set, data analysis using state-of-the-art NLP models, and a discussion of open challenges in this space.
The comprehensiveness of our data is limited by decisions about the data collection process, including which news outlets to focus on and which keywords and hashtags to use when collecting tweets.
While we take steps to broaden the coverage of our data, such as multiple rounds of identifying news outlets and relevant terms, collection biases could reduce the reliability of any analyses conducting with this data.
Our data set cannot be considered to capture all relevant content from this time period. 

Throughout \Sref{sec:analysis} we focus on highlighting the limitations of current NLP approaches in this setting, and we refer to this section for details. We acknowledge that our discussing of challenges and limitations is itself limited by the discussion framework. Structuring our analysis using different social science theories could lead to different results.
We additionally focus on entirely text analysis and do not discuss limitations related to other types of media, such as images or video \citep{beskow2019social}.
Finally, our discussion of limitations is based on our choice of NLP methodology to use over our data. While we attempt to select state-of-the-art models for the most dominant NLP research paradigms for each media effect, other methods and paradigms that reduce our discussed limitations may exist.

\section{Ethical Considerations}

Given the ongoing war and the limitations on free speech in Russia, including the recently passed law that punishes spreading ``false information'' with up to 15 years in prison, it is possible that our data set contains content that could have physical and legal ramifications for individual users or media outlets.
Even in our initial data collection, some VK data was flagged as deleted by moderators.
We take several steps to mitigate the impact our work may have on the risk to individuals or media outlets.
All of the data collected in this work is publicly available and we do not make any attempt to uncover non-public data.
While we do include posts by general users on Twitter, we primarily focus on posts from media outlets and replies to them, where we can assume a lower expectation of privacy.
In order to preserve users' ability to delete content, we do not release any raw text data and instead only release post IDs, which other researchers can use to recollect raw data, if it has not been removed.
We further note that all data was collected in accordance with social media platforms' terms of service.

Throughout this work, we also avoid using specific examples from the data or referring to individual users. We encourage future work on this data to exercise similar caution, and we do not condone any research that attempts to deanonymize or profile users or identify narratives that could result in individuals being targeted.
We refer to \citet{Vitak2016} and \citet{Williams2017} for a more in-depth discussion of ethical considerations of research using social media data.

We also primarily focus on news content posted by Russian media outlets, which we suggest provides avenues for studying disinformation, because of prior work on Russian information manipulation strategies and because Russia is the aggressor in this conflict. However, we note that independent reports have also found evidence of misinformation perpetuating pro-Ukranian narratives.\footnote{\href{https://www.newsguardtech.com/special-reports/russian-disinformation-tracking-center/}{https://www.newsguardtech.com/special-reports/russian-disinformation-tracking-center/}}
More generally, the authors of this work are situated in the U.S. and our assumptions in this work (e.g. that Russia is the aggressor) reflect this context, but we note that this viewpoint is not universal.

\section{Acknowledgments}
We thank the anonymous reviewers for their feedback, as well as the Text As Data 2022 audience, especially Sarah Dreier. A.F. and J.M. gratefully acknowledge support from the Google PhD Fellowship. C.Y.P. gratefully acknowledges support from KFAS. Y.T.gratefully acknowledges support from NSF CAREER Grant No.~IIS2142739, the Alfred P. Sloan Foundation Fellowship, and the DARPA Grant under Contract No.~HR001120C0124. Any opinions, findings and conclusions or recommendations expressed in this material are those of the author(s) and do not necessarily state or reflect those of the United States Government or any agency thereof.

\bibliographystyle{acl_natbib}
\bibliography{main, anthology}

\clearpage
\appendix

\section{Search Terms for Tweet Collection}
\label{appendix:twitter-search-terms}

Table \ref{tab:num_search_terms} describes the number of total search terms we curated in each update.
We describe each update in the following paragraphs. 
\paragraph{Version 0}
An initial round of defining hashtags and keywords; we manually collected general keywords and hashtags including (1) entity names, e.g. \textit{Russia}, \textit{Ukraine}, names of cities from both Ukraine and Russia, (2) war-related terms, including \textit{war}, \textit{peace}, \textit{UkraineRussianWar}, \textit{RussianUkraineWar}, and (3) we include the same keyword phrases in Russian and Ukrainian languages.  After we sampled 1K tweets with these general keywords, we sorted all hashtags in the data sample to augment this initial seed list. 

\paragraph{Version 1}
In a manual analysis of an initial data sample, we identified additional frequently mentioned entities, e.g. additional cities in Ukraine, names of politicians in Ukraine and Russia, stance-bearing pro-Russia and pro-Ukraine hashtags (e.g., \textit{\#IstandwithRussia}, \textit{\#stopputin}), additional hashtags referring to the Second World War (\#\russian{нацизм}), pro-war and anti-war hashtags (\textit{\#StopWar},  \#\russian{правдаовойне}). We note that while the overall sentiment in tweets was bearing more solidarity with Ukraine, the set of hashtags and keywords is diverse, in terms of languages (English, Russian, Ukrainian), stance (pro-Russa, pro-Ukraine, pro-war, anti-war), and in addition it includes mentions of external entities involved (NATO, Belarus, USA). 

\paragraph{Version 2}
We pulled an additional sample of 5K tweets using terms from Version 1 for a manual analysis of missing seed terms and to obtain the ranking of the terms and keywords by their frequency in the sample. 

\paragraph{Version 3 (Final)}
We analyzed the tweets collected through the first 24 hours and sorted hashtags by frequency. We then manually annotated top 665 hashtags (until freq-150) and added to the list 81 most frequent conflict-related hashtags. 

\begin{table}[h]
\centering
\begin{tabular}{c|cccc}
\textbf{Versions}&V0 & V1 & V2 & V3 (final)\\\hline
\textbf{\# of keywords} & 87 &98&184&264\\
\end{tabular}
\caption{The size of search term list in each update.}
\label{tab:num_search_terms}
\end{table}

\begin{tcolorbox}[colframe=white!75!black,width=0.95\linewidth,center]
\textbf{Pro-War Search Terms (8):}\\
\selectlanguage{russian}
\#crimeanspring, \#мненестыдно, \#русскиеидут, \#deadrussiansoldiers, \#istandwithputin, \#imwithrussia, \#своихнебросаем, \#proudtoberussian\\
\textbf{Anti-War Search Terms (12):}\\
\#нетвойне, \#stoprussia, \#nowar, \#stopputinnow, \#stopwarinukraine, \#nowarwithukraine,\#saveukraine, \#stopputin, \#нетвойнесукраиной, \#stopthewar, \#stopwar, \#stoprussianaggression
\selectlanguage{english}
\end{tcolorbox}

\begin{tcolorbox}[colframe=white!75!black,width=0.95\linewidth,center]
\textbf{Pro-Ukraine (19):}\\
\selectlanguage{russian}
\#standwithukraine, \#fkputin, \#нетпутину, \#slavaukraini, \#stoprussia, \#saveukraine, \#fklukashenko, \#stoprussianaggression, \#staywithukraine, \#своихнебросаем, \#stopputinnow, \#stopwarinukraine, \#nowarwithukraine, \#helpukraine, \#stopputin, \#banrussiafromswift, \#славаукраїні, \#istandwithukraine, \#istandwithzelenskyy\\
\selectlanguage{english}
\textbf{Pro-Russia (8):}\\
\selectlanguage{russian}
\#donbasstragedy, \#crimeanspring, \#русскиеидут, \#istandwithputin, \#istandwithrussia, \#imwithrussia, \#своихнебросаем, \#proudtoberussian
\selectlanguage{english}
\end{tcolorbox}

\textbf{Final Search Terms (264):}\\
\selectlanguage{russian}
\#Украина, \#нетвойне, \#Ukraine, \#Россия, \#украина, \#Харьков, \#НетВойне, \#Херсон, \#мариуполь, \#война, \#НетПутину, \#ukraine, \#mariupol, \#україна, \#Україна, \#StopWar, \#UkraineWar, \#нетвойнесУкраиной, \#Russia, \#россия, \#Путин, \#StopTheWar, \#путин, \#StopRussianAggression, \#Киев, \#Мариуполь, \#NoWarWithUkraine, \#UkraineRussie, \#StandWithUkraine, \#Зеленский, \#РФ, \#RussiaUkraineConflict, \#SaveUkraine, \#StopRussia, \#Сумы, \#UkraineInvasion, \#stopputin, \#СлаваУкраїні, \#UkraineRussiaCrisis, \#Гостомель, \#UkraineConflict, \#FlyAway, \#войска, \#ДНР, \#NoWar, \#Одесса, \#Харків, \#Kiev, \#Ukraina, \#России, \#Херсоне, \#путинубийца, \#протесты, \#Donbass, \#нацизм, \#Одеса, \#геноцид, \#Mariupol, \#eu, \#europe, \#фашизм, \#Odesa, \#Odessa, \#ЛНР, \#лукашенко, \#Москва, \#IStandWithUkraine, \#Мелитополь, \#невойне,
\#протесты, \#нацизм, \#фашизм, \#геноцид, \#война, \#WWII, \#nuclearwar, \#санкции, бомбит, нацизм, войска, диверсионные, Удары, армия, пиздец, мир, мирные,
\#DearsForPeace, МыНеМолчим, \#newsua, \#newsru, \#НетвойнеУкраиныпротивДонбасса, \#санктпетербург, \#зеленський, \#ДаПобеде, \#SWIFT, \#київ, \#мынемолчим, \#тихийпикет, \#ЕС, \#russianinvasion, \#Противійни, \#ПУТИН\_ВИНОВЕН, \#донбасс, \#EuroMaidan, \#Ирпень, \#беларусь, \#Maidan, \#МойЛуганск, \#StayWithUkraine, \#Zelenskiy, \#НетБезумию, \#питер, \#CoupdEtat, \#Протесты, \#бандеровцы, \#всу, \#Кремль, \#BanRussiafromSwift, \#бомбардировки, \#Лавров, \#Rusya, \#МОСКВА, \#АрмияРоссии, \#SanctionRussiaN, \#российское\_вторжение, \#ДавайЗаМир, \#НоваяКаховка, \#Irpin, \#worldwar3, \#Moscow, \#дапобеде, \#переговоры, \#русские, \#ООН, \#Евросоюз, \#путинхуйло, \#терроризм, \#Минобороны, \#WWIII, \#митинг, \#РусскаяВесна, \#DonbassWar, \#янемолчу, \#moscow, \#РоссияУбивает, \#русскийсолдат, \#времяпомогать, \#Шойгу, \#россияне, \#ЗаПрезидента, \#армия, \#наДонбассевойна8лет, \#МнеНеСтыдно, \#русскиймир, \#россияукраина, \#ЯМыПутин, \#ЕдинаяРоссия, \#DeadRussianSoldiers, \#ВКСРоссии, \#КремлевскиеСМИ, \#Русскиелюди, \#КризиснаДонбассе, \#денацификация, \#Putler, \#русскийТопот, \#россиявставай,
Путин, Россия, Украина, Киев, Путину, Украины, Россияне, АЭС, США, НАТО, Зеленский, \#Chernihiv, \#Kherson, \#Украина, \#Ukraine, \#Россия, \#украина, \#Харьков, \#Херсон, \#мариуполь, \#Киев, \#Мариуполь, \#Зеленский, \#РФ, \#Russia, \#россия, \#Путин, \#путин, \#ДНР, \#Харків, \#Kiev, \#России, \#Ukraina, \#Херсоне, \#донецк, \#Луганск
\#СвоихНеБросаем, \#DonbassTragedy, \#See4Yourself, \#Think4Yourself, \#WeRemember, \#IstandwithRussia, \#Novorossiya, \#Donbass, \#РаботайтеБратья, \#Welcome2Crimea, \#Crimea, \#CrimeanSpring, \#IStandWithPutin, \#своихнебросаем, \#русскиеидут, \#imwithrussia, \#ProudToBeRussian,
\#нетвойнесУкраиной, \#StopTheWar, \#StopRussianAggression, \#NoWarWithUkraine, \#StandWithUkraine, \#UkraineRussie, \#SaveUkraine, \#StopRussia, \#СлаваУкраїні, \#UkraineRussiaCrisis, \#UkraineInvasion, \#stopputin, \#NoWar, \#путинубийца, \#RussiaUkraineConflict, \#UkraineConflict, \#StopPutinNow, \#StopWar, \#StandWithUkraine, \#SlavaUkraini, \#HelpUkraine, \#invaision, \#РоссияБЕЗпутина, \#PutinIsFalling, \#PutinWarCrimes, \#StopWarInUkraine, \#resist, \#SlavaUkrayini, \#FreeBelarus, \#FKPutin, \#FKLukashenko, \#UkraineInvasion, \#правдаовойне, \#IStandWithZelenskyy, \#IStandWithUkraine, \#StopWarInUkraine, \#PutinWarCriminal, \#ClosetheSkyoverUkraine, \#AdolfPutin, \#PutinHitler, \#RussiaInvadedUkraine, \#нетвойне, \#НетВойне, \#НетПутину, \#UkraineWar
\selectlanguage{english}

\section{Twitter/VK Handles of Russian News Outlets}
\label{appendix:media-handles}

\begin{table}[h]
\centering
\resizebox{\columnwidth}{!}{%
\begin{tabular}{lll}
\textbf{Media Name}                    & \textbf{Twitter Handle} & \textbf{VK Handle} \\\hline
TV Rain                                & @tvrain                 & tvrain             \\
Alexei Navalny                         & @navalny                & navalny            \\
IStories                               & @istories\_media        & istories.media     \\
OVD-Info                               & @OvdInfo                & ovdinfo            \\
Novaya Gazeta                          & @novaya\_gazeta         & novgaz             \\
DW (Deutsche Welle)                    & @dw\_russian            &                    \\
BBC Russia                             & @bbcrussian             & bbc                \\
MediaZona & @mediazzzona            & mediazzzona        \\
Radio Liberty                          & @SvobodaRadio           & svobodaradio       \\
The Insider                            & @the\_ins\_ru           & theinsiders        \\
Forbes Russia                          & @ForbesRussia           & forbes             \\
Meduza                                 & @meduzaproject          & meduzaproject      \\
Current Time TV                        & @CurrentTimeTv          & currenttimetv      \\
RTVI                                   & @RTVi                   & rtvi               \\
Voice of America                       & @GolosAmeriki           & golosameriki       \\
Snob Project                           & @snob\_project          & snob\_project     \\
Echo of Moscow                         & @EchoMskRu              &                    \\
FBK       & @fbkinfo                &                    \\
Reuters Russia                         & @reuters\_russia        &                    \\
Znak.com  & @znak\_com              &                    \\\hline
\end{tabular}
}
\caption{List of Independent media and their handles on Twitter and VK.}
\end{table}

\begin{table*}[t]
\centering
\resizebox{.6\textwidth}{!}{%
\begin{tabular}{lll}
\textbf{Media Name}                           & \textbf{Twitter Handle} & \textbf{VK Handle}    \\\hline
RT (Russian)                                  & @RT\_russian            & rt\_russian                      \\
RT (English)                                  & @RT\_com                &            \\
TASS (Russian)                                        & @tass\_agency           & tassagency \\
TASS (English)                                & @tassagency\_en         &             \\
Sputnik News                                  & @SputnikInt             & sputnikint            \\
Sputnik (Radio)                               & @ru\_radiosputnik       & sputnik\_radio        \\
RIA Novosti                                   & @rianru                 & ria                   \\
RIA Novosti (Breaking News)                   & @riabreakingnews        &                       \\
PRIME                                         & @1prime\_ru             & 1prime                \\
Ministry of Defence & @mod\_russia            & mil                   \\
Ruptly                                        & @Ruptly                 & ruptly                \\
Moscow 24                                     & @infomoscow24           & m24                   \\
inoSMI                                        & @inosmi                 & inosmi                \\
Life                                          & @lifenews\_ru           & life                  \\
5TV                                           & @5tv                    & tv5                   \\
Vesti                                         & @vesti\_news            & vesti                 \\
Russia-1                                      & @tvrussia1              & russiatv              \\
RBC                                           & @ru\_rbc                & rbc                   \\
Gazeta.Ru        & @GazetaRu               & gazeta                \\
Rossiyskaya Gazeta                            & @rgrus                  & rgru \\
Ukraina.ru       & @ukraina\_ru            & ukraina\_ru\_official \\
Redfish                                       & @redfishstream          &                       \\
MIA Rossiya Segodnya                          & @pressmia               &                       \\
Margarita Simonyan                            & @M\_Simonyan            &                       \\
Zubovski 4                                    & @zubovski4              &                       \\
DVostok                                       & @media\_dv              &                       \\
Vladimir Soloviev                             & @VRSoloviev             &                       \\\hline
\end{tabular}}
\selectlanguage{english}
\caption{List of State-affiliated media and their handles on Twitter and VK.}
\label{tab:media-accounts}
\end{table*}


\begin{table*}[!htbp]
\centering
\resizebox{0.8\linewidth}{!}{%
\begin{tabular}{l|cc|cc|cc|c}
&\multicolumn{4}{c|}{\textbf{VK}} & \multicolumn{3}{c}{\textbf{Twitter}}\\
                           & \multicolumn{2}{c|}{\textbf{Media}} & \multicolumn{2}{c|}{\textbf{Public}}& \multicolumn{2}{c|}{\textbf{Media}} & \textbf{Public}  \\
                           & SA & Ind & SA    & Ind & SA    & Ind  &  \textit{Twit.}\\\hline
\textbf{Posts per account} &      26K             &     11K       &          34.7           &    59.8 &5.6K                   &5.8K             &23.1    \\

\textbf{Word count}       & 26.1              & 50.8        & 14.9                & 16.7      & 19.2              & 23.3        & 19.2      \\\hline
\textbf{Image/video (\%)}    & 70.2              & 21.4        & 8.2                 & 12.3     & 50.9              & 28.0       & 9.6         \\
\textbf{Link (\%)}             & 26.5              & 76.3        & 0.2                 & 0.9        & 78.8              & 75.3        & 7.1       \\\hline
\textbf{Likes}                & 81.9              & 66.8        & 2.0                 & 2.3    & 39.4              & 249.4        & 4.3            \\
\textbf{Comments/RTs}             & 39.3              & 25.9        & -                   & -    & 11.2              & 60.0        & 399.7           \\
\textbf{Views}                & 18K           & 10K     & -                   & -      &-&-& -      \\
\end{tabular}
}
\caption{Statistics of \dataname. The difference between state-affiliated and independent media was statistically significant ($p<0.05$) for all metrics in both Media Posts and Public Reaction.}
\label{tab:vk_stats}
\end{table*}

\section{Data Statistics and Analysis}
\label{appendix:data-stats}

\subsection{Analysis: Post Content}
\paragraph{Length} As a consequence of Twitter's 280-character constraint, VK posts are on average significantly longer than Twitter posts. Interestingly, independent media posts are significantly longer than state-affiliated media posts on both platforms. This pattern is consistent in comments on VK.

Images and videos can themselves be powerful framing devices \citep{powell2015clearer}, and images posted to VK in particular have been used to understand opposing representations and interpretations of the Russia-Ukraine conflict \citep{makhortykh2017social}. On both VK and Twitter, state-affiliated media posts include much more multimedia (images and video) than independent media posts (70.3\% vs. 21.5\% on VK and 59.6\% vs. 31.9\% on Twitter, respectively).

\paragraph{External links}
In contrast to embedded multimedia, a slightly different pattern emerges for the inclusion of external links, which have been shown to enhance users' perceptions of trustworthiness and credibility on social media \citep{morris2012tweeting,wang2013trust}. 
The majority of both state-affiliated and independent media posts on Twitter include external links (70.5\% and 72.5\%, respectively), possibly again a consequence of Twitter's constraint affordance. However, there is a stark difference on VK, where 76.3\% of independent media posts include external links compared to just 26.4\% of state-affiliated posts.  As expected, a much lower proportion of public Tweets and public comments to media posts on VK contain embedded multimedia or external links. 
Public tweets collected via hashtags have slightly higher rates of including multimedia and links compared to VK comments, but are much lower compared to media posts (9.9\% of public tweets include images or video, and 6.9\% include external URLs). 

\subsection{Analysis: Activity and User Engagement}
Analyses of account activity and user engagement suggests that state-affiliated media dominates VK, but independent media dominates Twitter. 

\paragraph{Account Activity} On average, each state-affiliated media account included in \datanamevk{} has nearly 25K posts, more than twice as much as independent media which averages 11K posts per account. This pattern is reversed on Twitter, where independent accounts are slightly more active than state-affiliated accounts. 

We also observe a high degree of self-sorting among users who comment on VK media posts: 74.4\% comment only on state-affiliated posts, 16.8\% only on independent posts, and only 8.8\% of users have commented on both types of media posts. In other words, most people who comment on state-affiliated posts never comment on independent posts and vice versa. While we do not have user-level data about media exposure, this pattern suggests that information from state-affiliated and independent media reach disparate audiences.

\paragraph{Views} Unlike most platforms studied by NLP and computational social science researchers, VK's publicly-available data includes view counts (i.e. impressions) and thus presents a unique opportunity to study incidental exposure to media content \citep{tewksbury2001accidentally}. Not only are state-affiliated outlets more active on VK than independent outlets, but also each post on average reaches a larger audience (17K vs 10K views, respectively).

\paragraph{Interactive engagement metrics} 
Popularity cues, such as the numbers of likes, comments, and retweets, can serve as an indicator of the success of the media's agenda-setting, framing, and propaganda strategies. These popularity cues have further consequences: they can be used to recommend content on social media platforms and thus impact users' media diets, and they can act as heuristics for people trying to decide what media content is credible, accurate, and important \citep{haim2018popularity,porten2018popularity}. Consistent with the idea that the Twitter public sphere is more globally-oriented, independent media posts receive more engagement on Twitter than state-affiliated posts. In contrast, state-affiliated media posts on VK receive more engagement than independent posts. However, we note that independent posts on VK still have a higher rate of engagement if we account for their smaller audiences (view counts).

\subsection{Volume over Time}
\label{appendix:twitter-volume}

\begin{figure*}[h]
  \centering
  \subfloat{\label{fig:vk_vol_post}\includegraphics[width=0.49\textwidth]{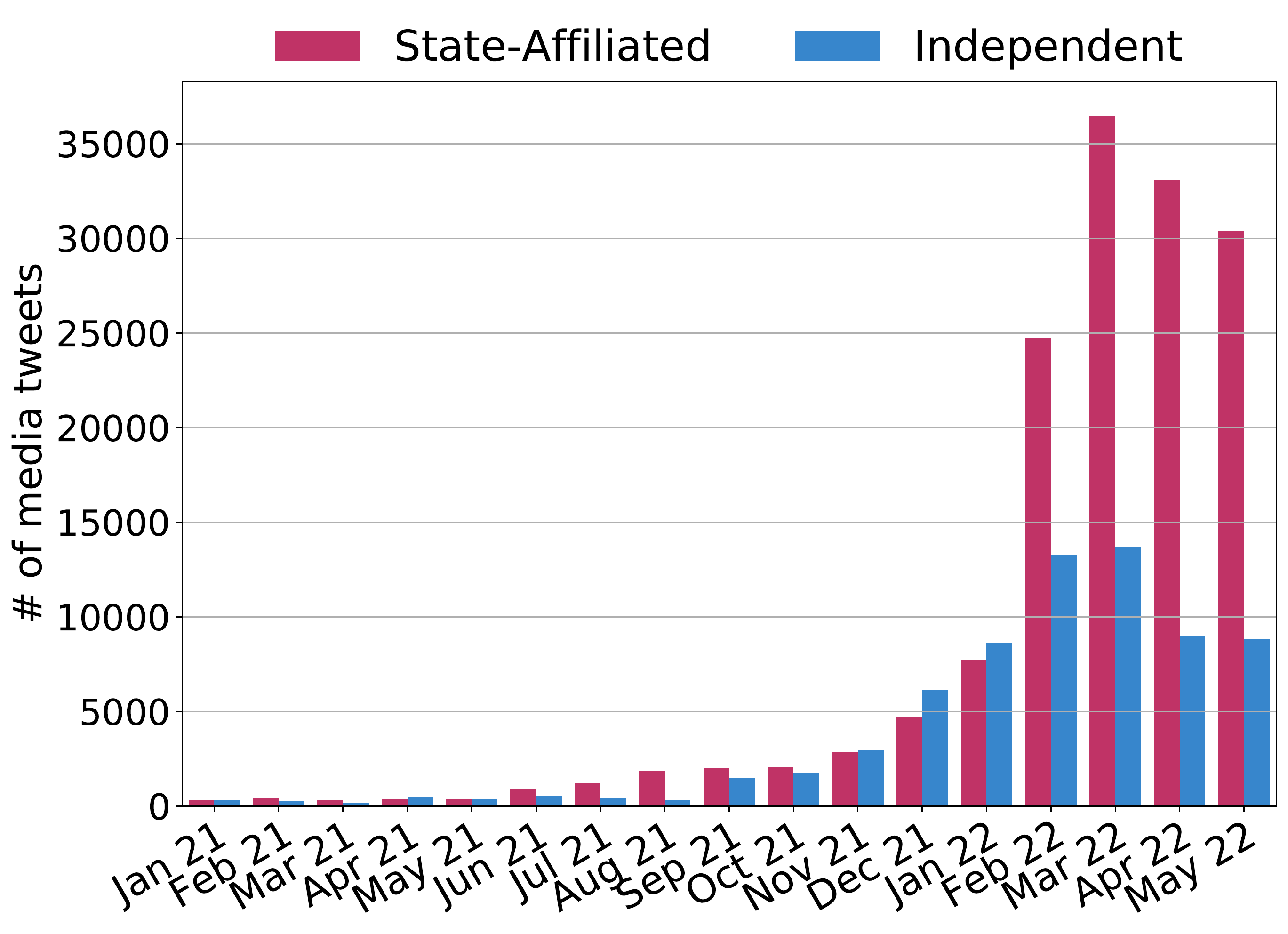}}
  \subfloat{\label{fig:vk_vol_comment}\includegraphics[width=0.49\textwidth]{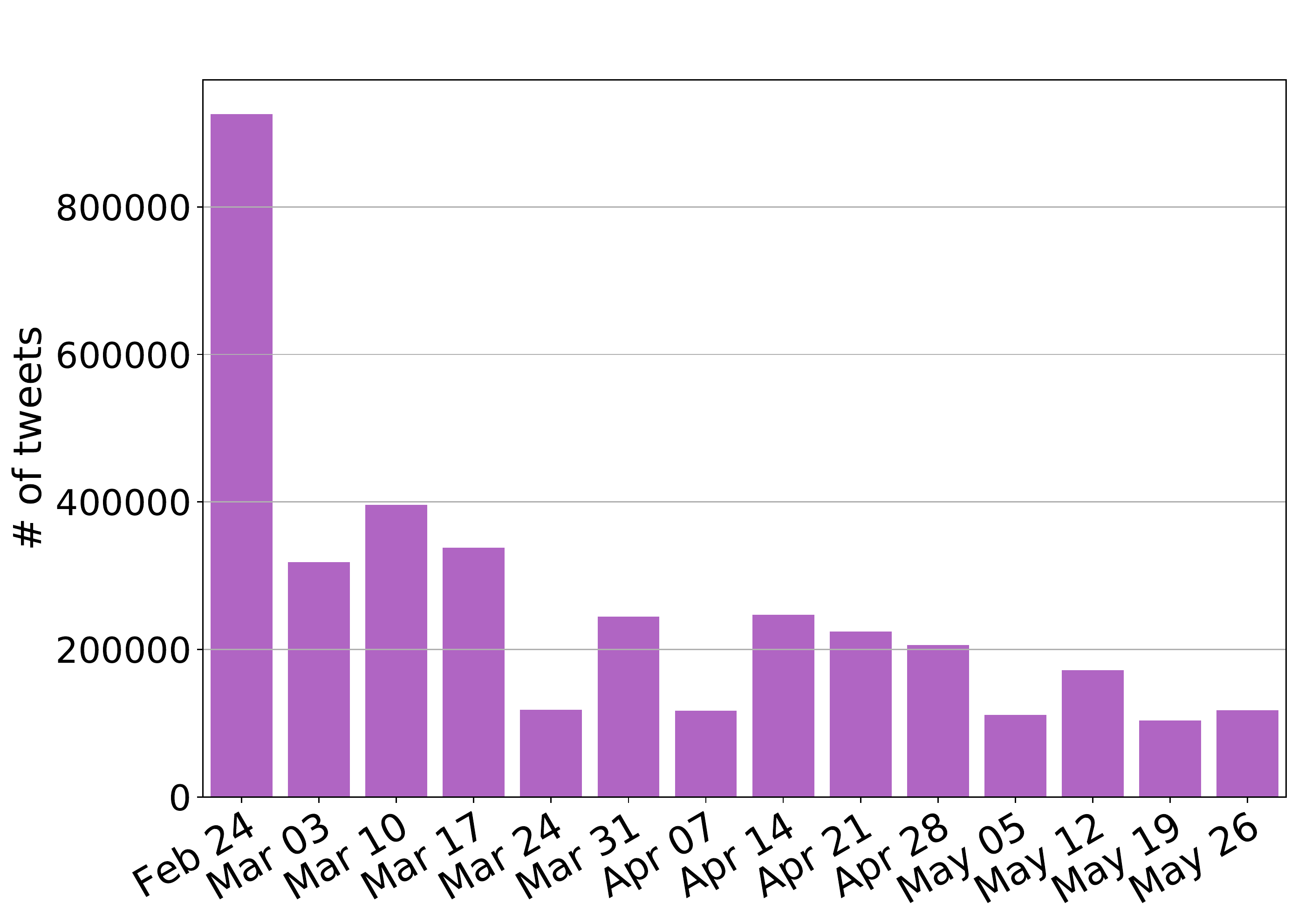}}
  \caption{Total volume of tweets from media accounts (left) and general accounts (right) in \datanametwitter.}
  \label{fig:vol_twitter}
\end{figure*}

\begin{figure*}[h]
  \centering
  \subfloat{\label{fig:vk_vol_post}\includegraphics[width=0.49\textwidth]{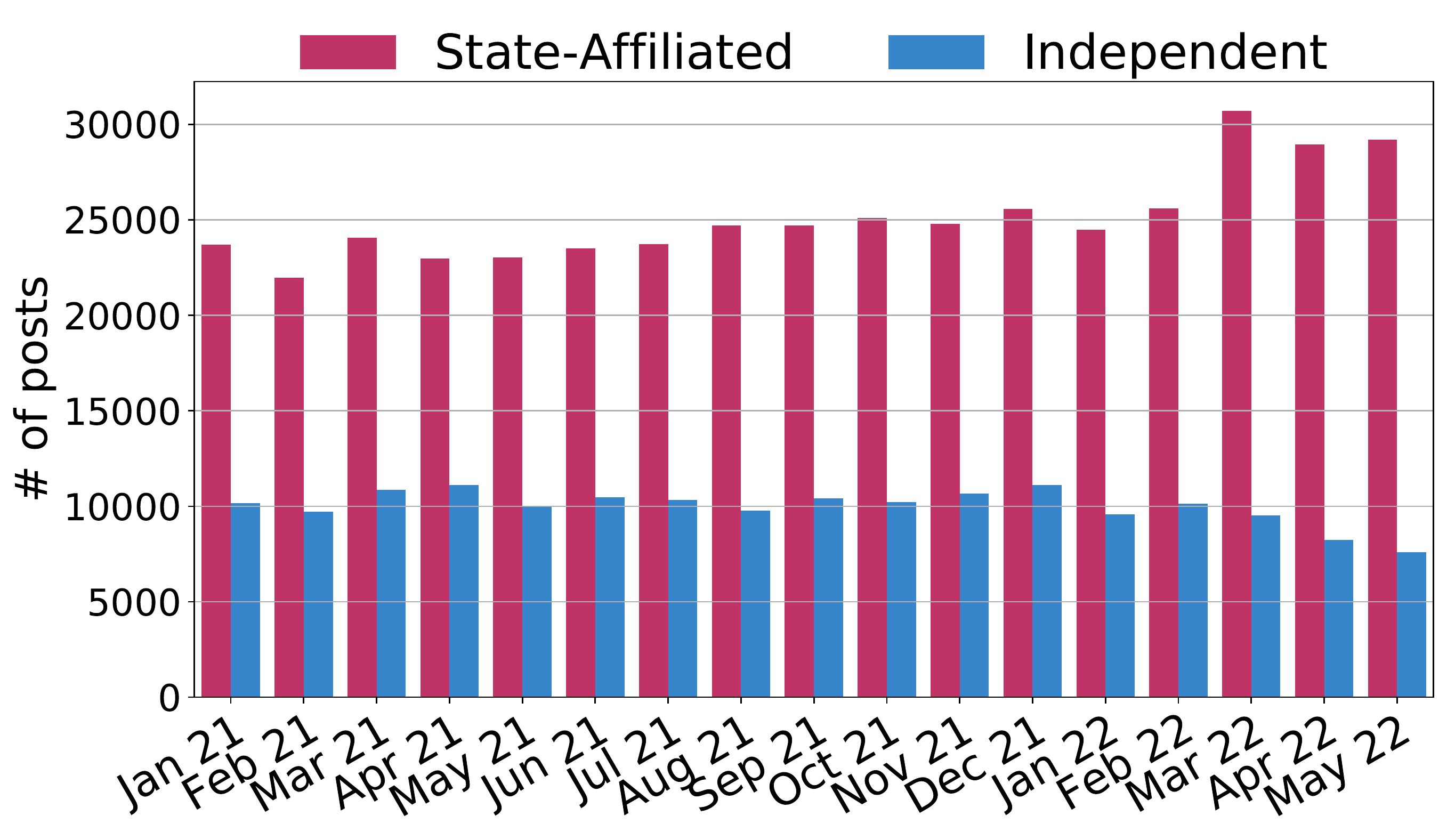}}
  \subfloat{\label{fig:vk_vol_comment}\includegraphics[width=0.49\textwidth]{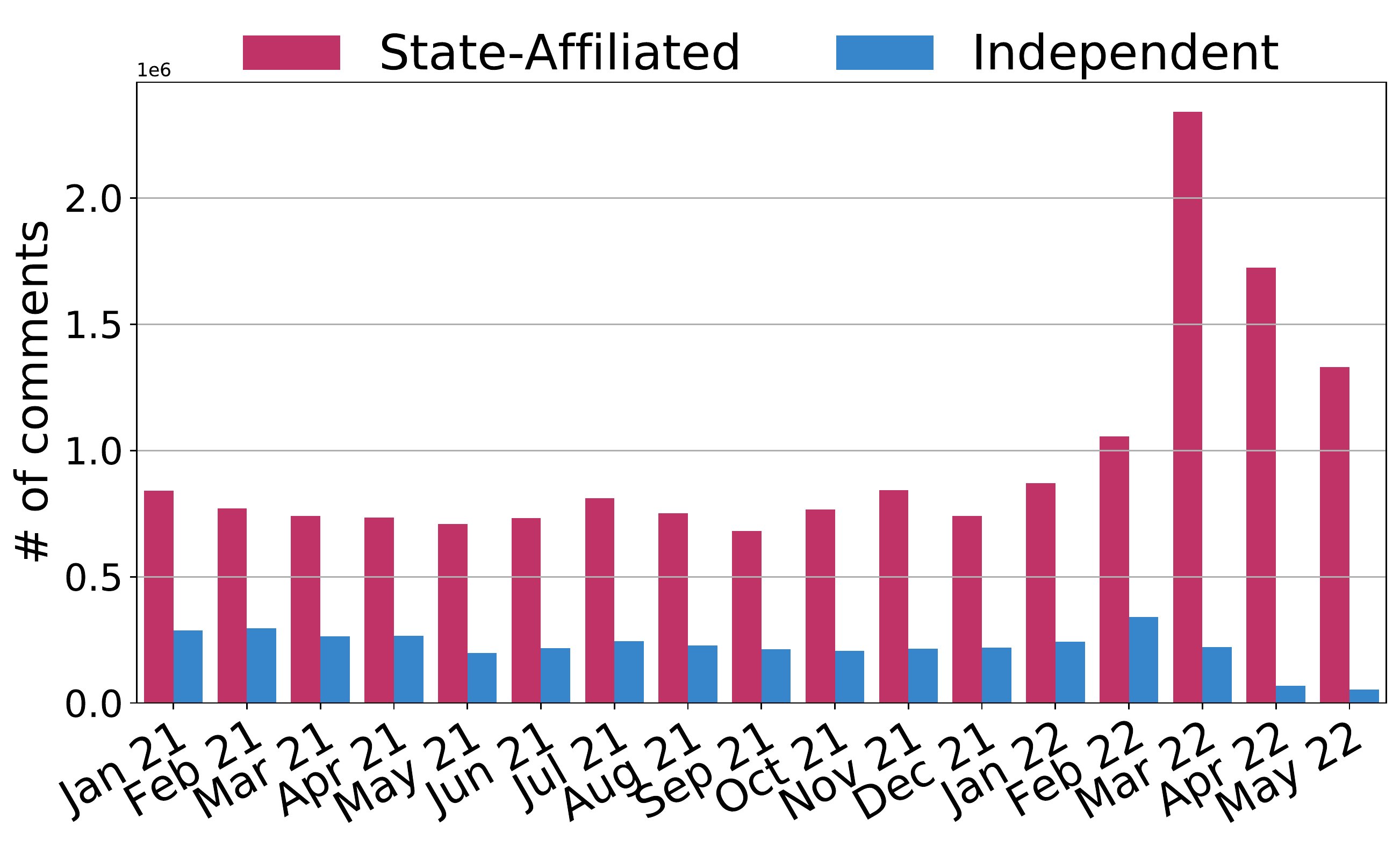}}
  \caption{Total volume by month of posts (left) and comments (right) in \datanamevk.}
  \label{fig:vol_vk}
\end{figure*}

\begin{figure*}[!htbp]
  \centering
  \subfloat{\label{fig:vk_vol_post}\includegraphics[width=0.49\textwidth]{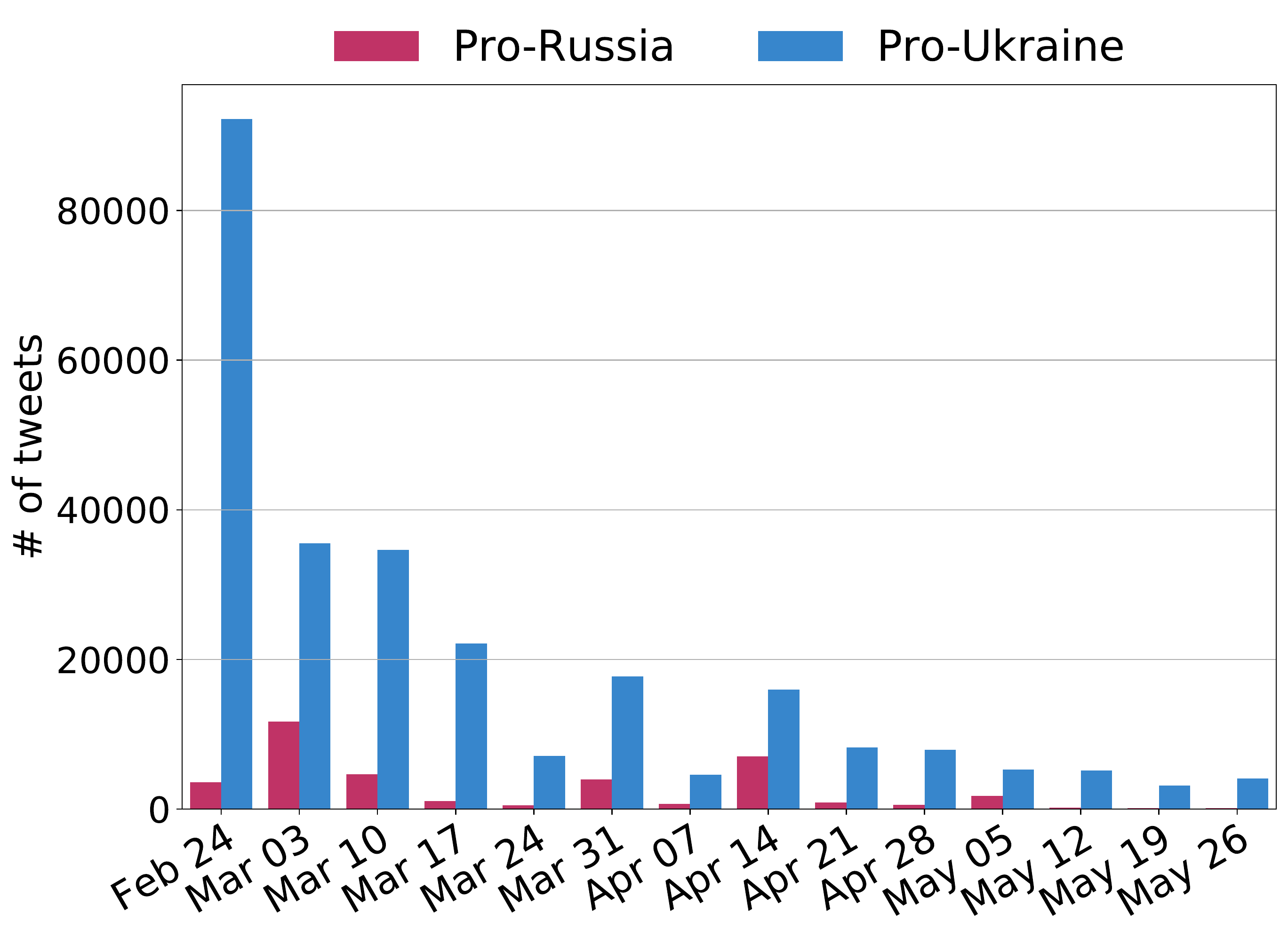}}
  \subfloat{\label{fig:vk_vol_comment}\includegraphics[width=0.49\textwidth]{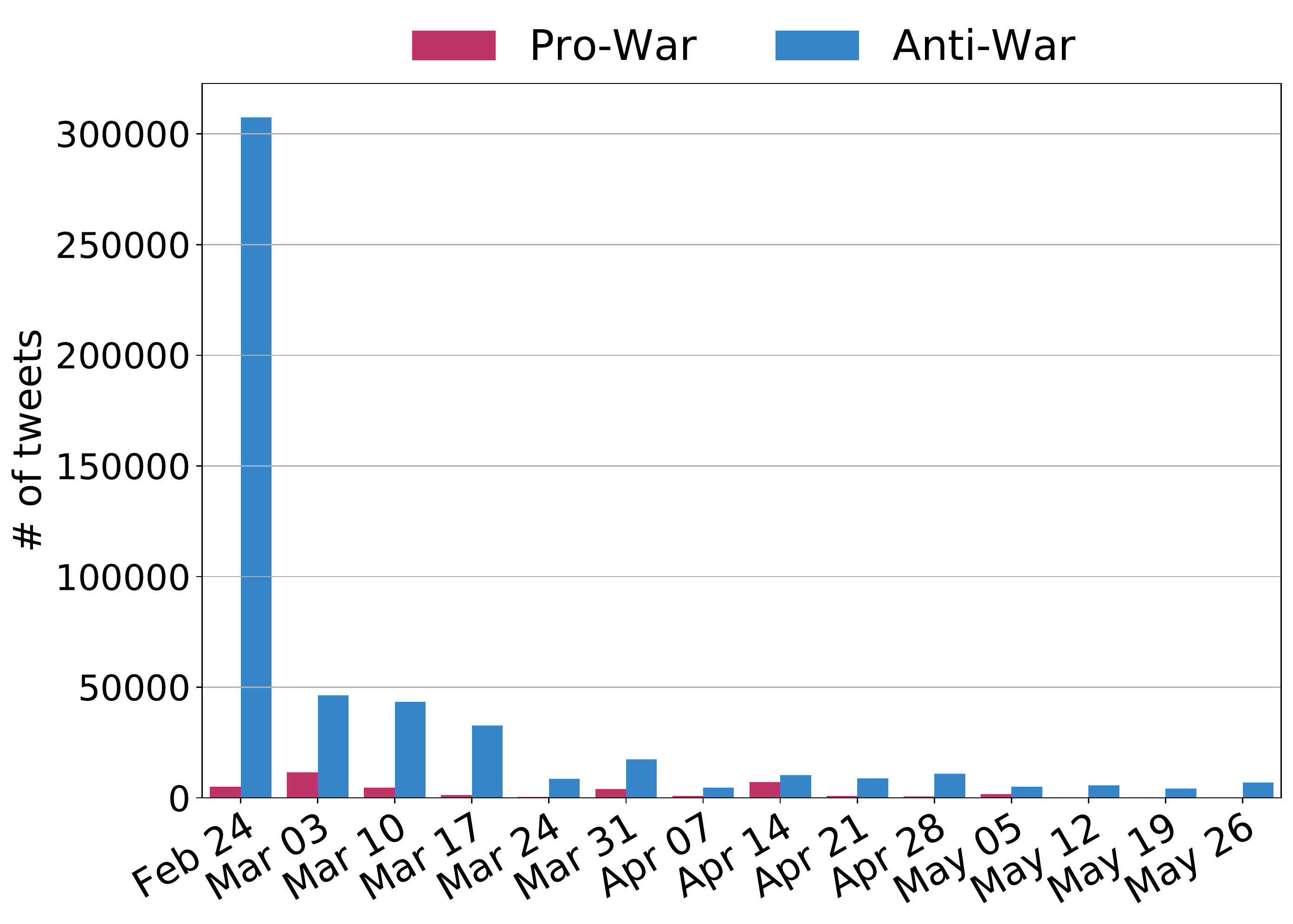}}
  \caption{Weekly volume of tweets that contain Pro-Russia/Pro-Ukraine (left) and Pro-War/Anti-War (right) hashtags during the war.}
  \label{fig:stance_twitter}
\end{figure*}

In February and March 2022, immediately after the war began, the volume of posts by media accounts and comments in both \datanamevk{} and \datanametwitter{} significantly increased (\Fref{fig:vol_vk} and \Fref{fig:vol_twitter}).
However, on March 4, Putin signed a new bill called ``fake news laws'' which punishes spreading ``false information'' with up to 15 years in prison. Consequently, many independent Russian media outlets including TV Rain and Radio Liberty temporarily suspended operations, while others announced that they were stopping coverage of the invasion because of the signed bill; these independent outlets include Colta.ru, Snob Project, Znak.com, and Novaya Gazeta.\footnote{\href{https://www.amnesty.org/en/latest/news/2022/03/russia-kremlins-ruthless-crackdown-stifles-independent-journalism-and-anti-war-movement/}{https://www.amnesty.org/en/latest/news/2022/03/russia-kremlins-ruthless-crackdown-stifles-independent-journalism-and-anti-war-movement/}}
The impact of the censorship is also evident in our data set, as we see a significant decrease in the volume of independent media accounts' posts and comments to independent media starting March 2022.

We also note that state-affiliated media accounts became extremely active on Twitter after the war started, even when compared to their own activity on VK. For instance, the number of state-affiliated tweets in the first half of May greatly surpasses the volume from the first half of April, but the opposite trend is observed on VK. This suggests a recent shift in Russia's state-affiliated media strategy: they are focusing more efforts on reaching and spreading (dis)information to a global audience through Twitter, rather than a primarily Russian audience through VK.

While we can divide media posts and their comments according to state-affiliated and independent outlets, we do not have user-level information about individual Twitter users' stances towards Russia or the war.  However, among the search terms we used for data collection, we curated a list of terms that show clear association with certain stances (Pro-Russia/Pro-Ukraine/Pro-War/Anti-War), which can be found in \Aref{appendix:twitter-search-terms}.
We then measure the volume of tweets that contain such stance-related terms (\Fref{fig:stance_twitter}).
The results show that Pro-Ukraine and Anti-war tweets are consistently more prominent on Twitter.
Russian-speaking Twitter users tend to be more pro-Ukraine and Anti-war compared to Russian residents according to a recent poll result that shows 81\% of Russian people support the Russian military operation in Ukraine.\footnote{\href{https://www.levada.ru/en/2022/04/11/the-conflict-with-ukraine/}{https://www.levada.ru/en/2022/04/11/the-conflict-with-ukraine/}} 
Considering the fact that VK is more widely used inside of Russia than Twitter, our results suggest researchers should exercise caution in generalizing opinions on Twitter to the entire Russian population.

\begin{figure}[!ht]
    \centering
    \includegraphics[width=0.5\textwidth]{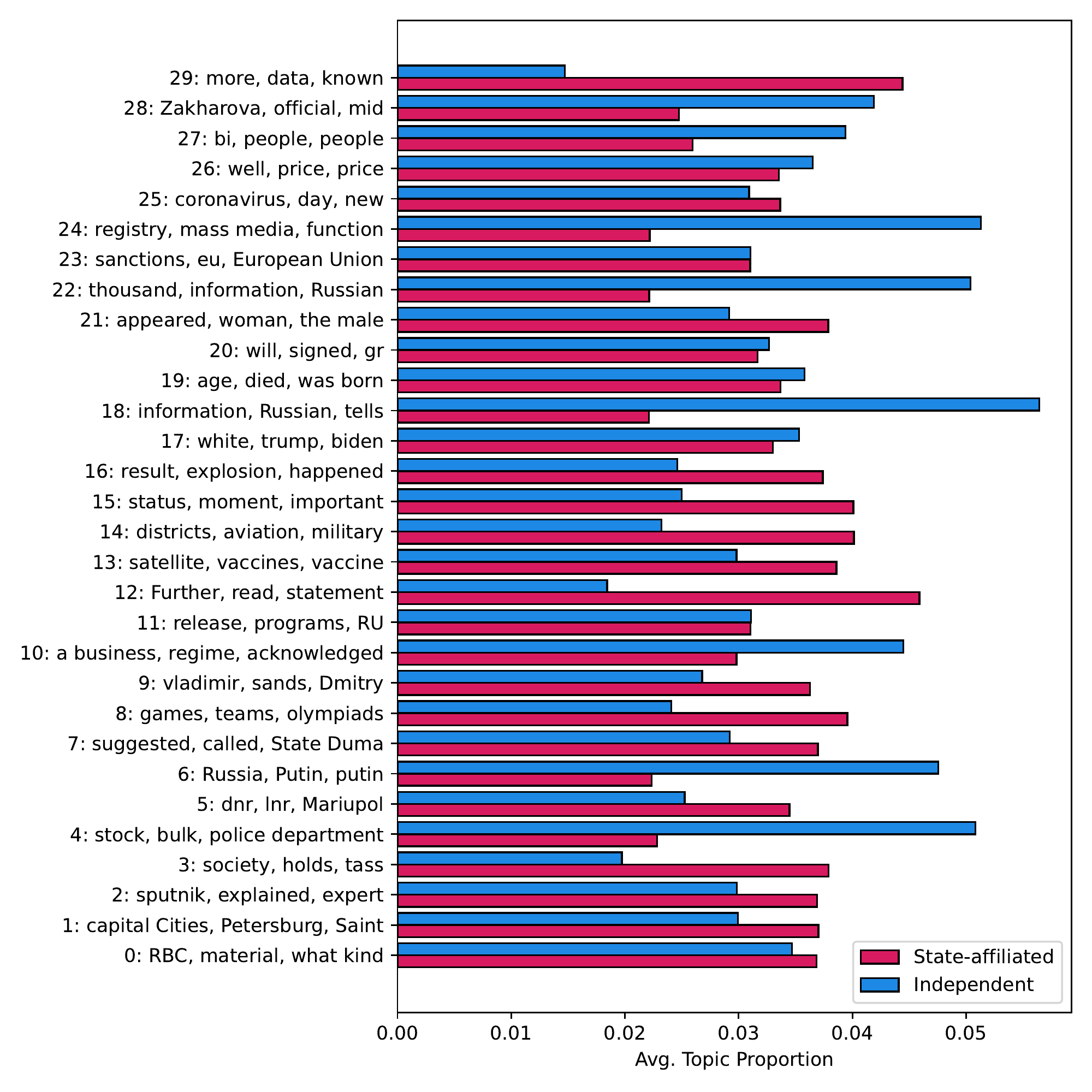}
    \caption{Topic proportions for state-affiliated and independent media outlets, as learned by a 30-topic CTM. The y-axis lists the highest-probable words for each learned topic. The x-axis reflects the estimated topic proportion, averaged across all state-affiliated or independent news outlets.}
    \label{fig:neural_all}
\end{figure}

\begin{figure}[!ht]
    \centering
    \includegraphics[width=0.4\textwidth]{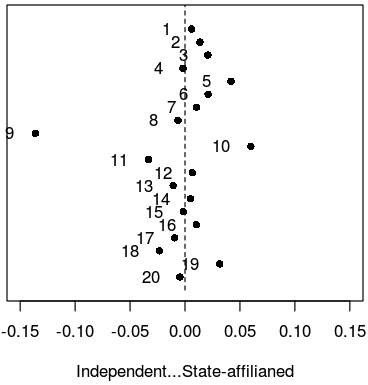}
    \caption{Topic proportion shifts between independent and state-affiliated outlets estimated by a 20-topic STM model, where time and outlet affiliation are incorporated in the model as topic prevalence covariates.}
    \label{fig:stm_all}
\end{figure}

\begin{figure}[!ht]
    \centering
    \includegraphics[width=0.5\textwidth]{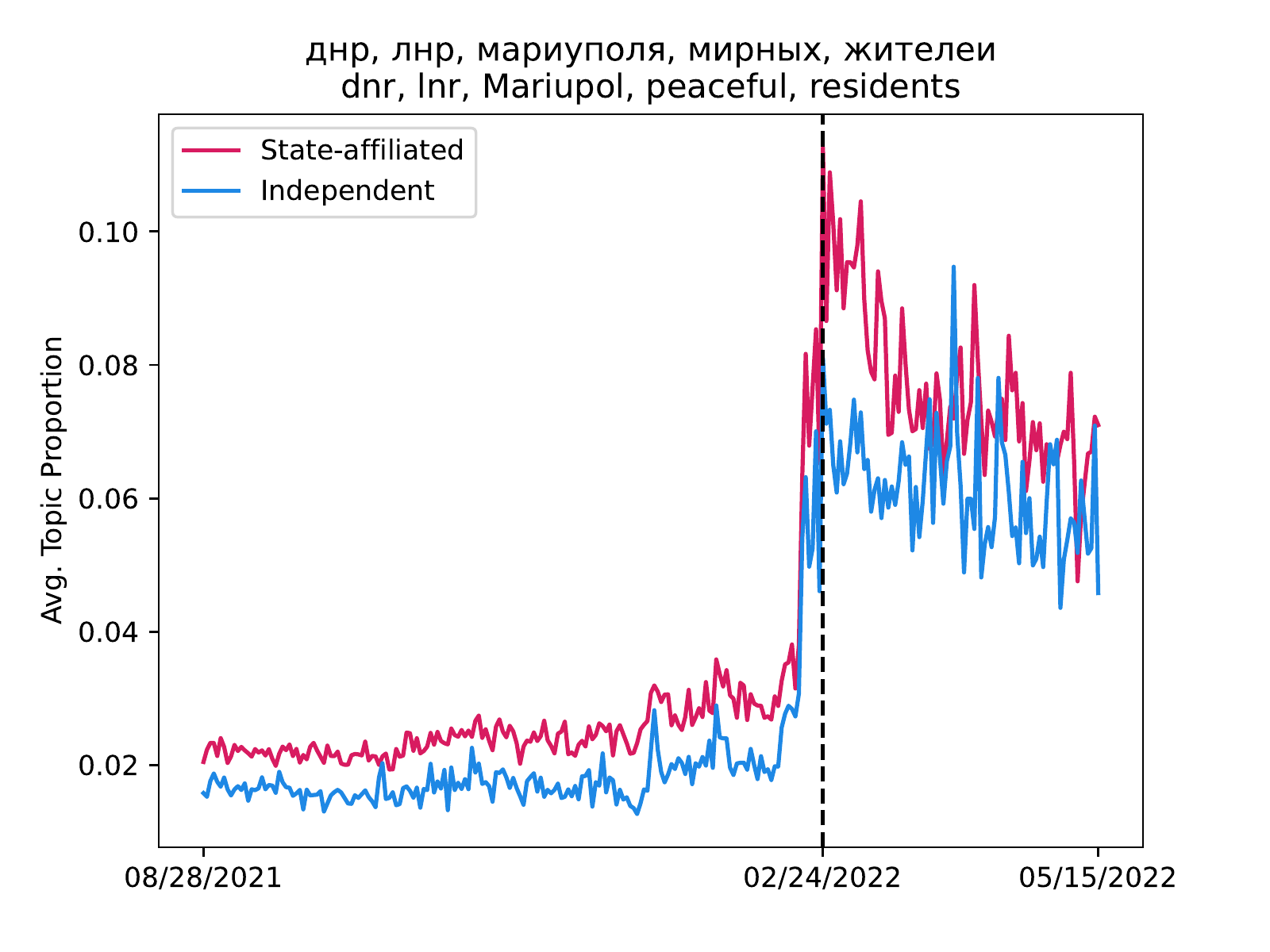}
    \caption{Average estimated proportion of documents related to Topic 5, learned by the CTM model. Topic 5 is the learned topic with the highest probability of Ukraine related terms. Estimated topic prevalence increases sharply in the days preceding the invasion.}
    \label{fig:neural_topic5_time}
\end{figure}

\section{Topic Model Parameters and Additional Data}
\label{appendix:topic_model_params}
We train the CTM model with 30 topics and take contextualized embeddings from \textsc{paraphrase-multilingual-mpnet-base-v2}.\footnote{\url{https://huggingface.co/sentence-transformers/paraphrase-multilingual-mpnet-base-v2}} Contextual embeddings are derived from the first 128 tokens in each document. We preprocess data by removing stopwords and words that occur in $>99.5\%$ of documents.
We train for 50 epochs.
We train the STM with 20 topics, and set news outlet affiliation and days since the corpus-collection start as topic-prevalence covariates (e.g., $prevalence =\sim kind * s(date)$). We train for 75 epochs using Spectral initialization. Both models are trained on data through May 15, 2022.
For both models, we fixed the number of topics based on which output looked most coherent out of 10, 20, and 30 topics.

\begin{figure}[!htbp]
    \centering
    \includegraphics[width=0.5\textwidth]{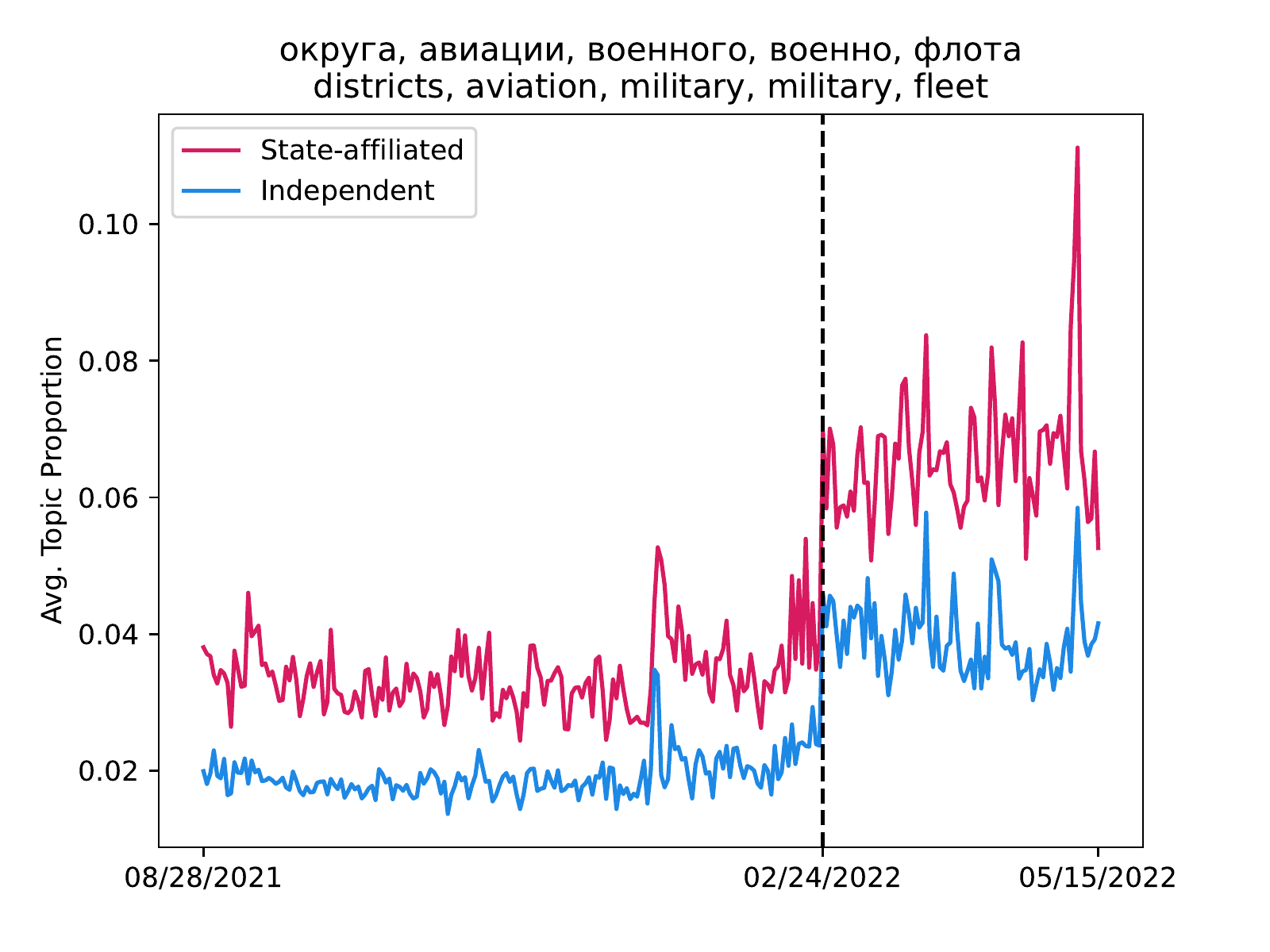}
    \caption{Average estimated proportion of documents related to Topic 14, learned by the CTM model. Topic 14 is the learned topic with the highest probability of military related terms. Estimated topic prevalence increases sharply in the days preceding the invasion.}
    \label{fig:neural_topic14_time}
\end{figure}

\begin{figure}[!htbp]
    \centering
    \includegraphics[width=0.4\textwidth]{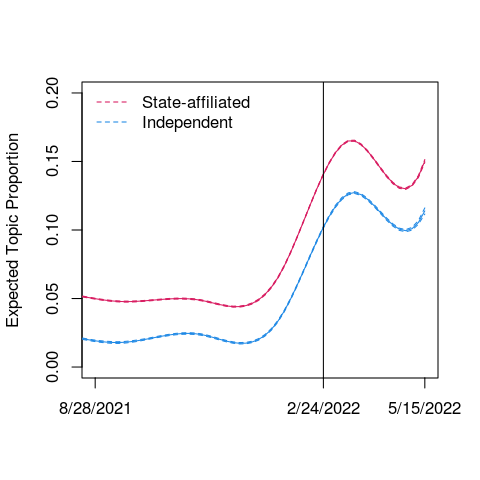}
    \caption{Expected topic proportions of Topic 19 learned by the STM model overtime. Topic 19 is the learned topic with the highest probability of Ukraine and military related terms (e.g., Ukraine, Russian, defense), and expected proportion of this topic increase sharply around the time of the invasion in both state-affiliated and independent news outlets.}
    \label{fig:stm_ukraine_over_time}
\end{figure}

\Tref{tab:neural_topics} lists the most probable words in each topic identified by the CTM model and \Tref{tab:stm_topics} lists the same for the STM model. Figures \ref{fig:neural_all} and \ref{fig:stm_all} show topic prevalence as associated with-affiliated or independent outlets.  \Fref{fig:neural_topic5_time}, \Fref{fig:neural_topic14_time}, and \Fref{fig:stm_ukraine_over_time} show topic prevalence over time for the topics most related to Ukraine and the war.

\begin{table*}[!ht]
    \centering
    \resizebox{\textwidth}{!}{
    \begin{tabular}{lll}
0 & \russian{рбк, материале, какие, life, рассказываем, разбираемся, читаите, forbes, бизнес, чаще} & RBC, material, what, life, we tell, understand, read, forbes, business, often \\
1 & \russian{столицы, петербурге, санкт, столице, москве, днем, утро, друзья, градусов, ожидается} & capital, petersburg, st, capital, moscow, afternoon, morning, friends, degrees, expected \\
2 & \russian{sputnik, объяснил, эксперт, радио, ру, би, рассказал, рассказала, интервью, си} & sputnik, explained, expert, radio, ru, bi, told, told, interview, si \\
3 & \russian{общество, проводит, тасс, экономика, ruptly, политика, reuters, премьер, мире, михаил} & society, conducts, tass, economy, ruptly, politics, reuters, premier, world, michael \\
4 & \russian{акции, навального, овд, инфо, протеста, поддержку, задержанных, алексея, акциях, новои} & actions, Navalny, OVD, info, protest, support, detainees, aleksey, actions, new \\
5 & \russian{днр, лнр, мариуполя, мирных, жителеи, украинские, народнои, новости, донбасса, мариуполе} & DPR, LPR, Mariupol, peaceful, residents, Ukrainian, folk, news, Donbass, Mariupol \\
6 & \russian{россии, путина, путин, это, функции, иностранного, агента, политолог, выполняющим, почему} & Russia, Putin, Putin, this, functions, foreign, agent, political scientist, doing, why \\
7 & \russian{предложили, назвали, госдума, предупредили, депутаты, предлагают, доступ, законопроект, новую, предлагает} & proposed, named, State Duma, warned, deputies, offer, access, draft law, new, offers \\
8 & \russian{игр, сборнои, олимпиады, команды, олимпиаде, спортсменов, чемпионата, золото, команда, победу} & games, teams, olympiads, teams, olympics, athletes, championship, gold, team, victory \\
9 & \russian{владимир, песков, дмитрии, президент, путин, зеленскии, секретарь, кремль, переговоров, лукашенко} & Vladimir, Sands, Dmitry, President, Putin, Zelensky, Secretary, Kremlin, negotiations, Lukashenko \\
10 & \russian{дело, режима, признал, приговор, свободы, годам, обвинение, лишения, бывшего, грозит} & case, regime, admitted, verdict, freedom, years, accusation, deprivation, former, threatens \\
11 & \russian{выпуск, программы, ru, смотрим, программе, шоу, utm, россия, эфир, телеканале} & release, programs, ru, watch, programme, show, utm, russia, air, TV channel \\
12 & \russian{далее, читать, заявление, сделал, важное, принял, сделали, сделала, приняли, обратился} & further, read, statement, did, important, accepted, did, did, accepted, applied \\
13 & \russian{спутник, вакцины, вакцину, коронавируса, вакцина, вакцин, вакцинации, воз, омикрон, здравоохранения} & satellite, vaccine, vaccine, coronavirus, vaccine, vaccine, vaccination, who, omicron, health \\
14 & \russian{округа, авиации, военного, военно, флота, учения, су, военнослужащие, противника, сил} & county, air, military, military, navy, exercise, su, servicemen, enemy, forces \\
15 & \russian{статус, момента, важных, продолжается, отказались, сказала, выход, собираются, временем, руководство} & status, moment, important, continues, refused, said, exit, going to, by the time, leadership \\
16 & \russian{результате, взрыв, произошел, борту, погиб, пострадавших, аварии, предварительным, находились, пожара} & result, explosion, occurred, board, killed, injured, accident, preliminary, were, fire \\
17 & \russian{белого, трампа, баиден, трамп, баидена, сша, джо, администрация, энтони, белыи} & white, trump, baiden, trump, baiden, usa, joe, administration, antony, whites \\
18 & \russian{информации, россииским, рассказывает, сообщение, массовои, материал, функции, иностранного, иностранным, агента} & information, Russian, tells, message, mass, material, functions, foreign, foreign, agent \\
19 & \russian{возрасте, умер, родился, скончался, жизни, ссср, рождения, актер, роли, советского} & age, died, born, deceased, life, ussr, birth, actor, roles, soviet \\
20 & \russian{будут, подписал, qr, смогут, выплаты, правительство, поддержки, закон, должны, могут} & will, signed, qr, may, payments, government, support, law, must, may \\
21 & \russian{появилось, женщина, мужчина, девушка, instagram, летняя, женщину, мать, ребенка, житель} & appeared, woman, man, girl, instagram, summer, woman, mother, child, inhabitant \\
22 & \russian{тысяч, информации, россииским, рублеи, сообщение, массовои, миллиона, около, материал, млн} & thousand, information, Russian, rubles, message, mass, million, about, material, million \\
23 & \russian{санкции, ес, евросоюз, против, евросоюза, отношении, пакет, россииских, запрет, дипломатов} & sanctions, eu, eu, against, eu, regarding, package, Russian, ban, diplomats \\
24 & \russian{реестр, сми, функцию, писали, выполняет, требует, иноагентов, нко, иноагента, закон} & registry, media, function, wrote, performs, requires, foreign agents, nco, foreign agent, law \\
25 & \russian{коронавирусом, сутки, новых, последние, число, случаев, умерли, случая, заболевших, максимум} & coronavirus, days, new, last, number, cases, dead, cases, cases, max \\
26 & \russian{курс, цена, стоимость, выросла, выросли, цены, вырос, tesla, рост, цен} & rate, price, value, up, up, prices, up, tesla, up, price \\
27 & \russian{би, людеи, люди, это, си, которые, воины, власти, несколько, время} & bi, people, people, this, si, which, warriors, powers, several, time \\
28 & \russian{захарова, официальныи, мид, мария, представитель, информации, россииским, сообщение, массовои, материал} & Zakharova, official, mid, maria, representative, information, Russian, message, mass, material \\
29 & \russian{подробнее, данные, известно, прокомментировали, обратился, ситуации, выступил, стало, оценили, отреагировали} & more, data, known, commented, applied, situations, speak, became, appreciated, reacted \\
    \end{tabular}
    }
    \caption{Highest probability words for each topic learned by 30-topic CTM model}
    \label{tab:neural_topics}
\end{table*}

\begin{table*}[!ht]
    \centering
    \resizebox{\textwidth}{!}{
    \begin{tabular}{lll}
1 & \russian{рассказал, далее, интервью, наш, андрей, главный, эксперт} & told, further, interview, our, andrey, chief, expert \\
2 & \russian{читать, своих, заявили, прокомментировал, слова, эксперт, возможность} & read, own, stated, commented, words, expert, opportunity \\
3 & \russian{россии, новости, ситуации, россию, заявление, насчёт, газа} & russia, news, situations, russia, statement, about, gas \\
4 & \russian{изза, ранее, сми, дело, данным, человека, сообщили} & because of, earlier, media, case, data, person, reported \\
5 & \russian{httpsliferup, видео, области, фото, смотрим, дома, тасс} & httpsliferup, video, areas, photo, look, houses, tass \\
6 & \russian{страны, глава, сообщил, сергей, мид, новые, стран} & countries, head, informed, sergey, mid, new, countries \\
7 & \russian{детей, стали, рассказали, результате, известно, погибли, одной} & children, became, told, result, known, died, one \\
8 & \russian{власти, суд, решение, москвы, связи, мая, такое} & authorities, court, decision, moscow, communications, may, such \\
9 & \russian{иностранного, агента, функции, выполняющим, информации, российским, иностранным} & foreign, agent, function, performer, information, Russian, foreign \\
10 & \russian{подробнее, заявил, россия, путин, владимир, президент, александр} & more, stated, russia, putin, vladimir, president, alexander \\
11 & \russian{рублей, тысяч, навального, новой, января, делу, акции} & rubles, thousands, bulk, new, january, deeds, shares \\
12 & \russian{россиян, около, coronavirus, мире, тыс, стране, дней} & Russians, around, coronavirus, world, thousand, country, days \\
13 & \russian{также, которые, могут, будут, компании, пока, которых} & also, which, can, will, companies, yet, which \\
14 & \russian{москве, дня, стало, февраля, апреля, рассказываем, нашем} & Moscow, days, became, February, April, we tell, our \\
15 & \russian{сша, против, президента, путина, считает, российского, безопасности} & usa, against, president, putin, believes, Russian, security \\
16 & \russian{коронавируса, covid, сутки, новых, последние, россии, коронавирусом} & coronavirus, covid, day, new, latest, russia, coronavirus \\
17 & \russian{года, лет, году, день, несколько, жизни, год} & years, years, year, day, multiple, life, year \\
18 & \russian{это, почему, людей, люди, очень, рассказывает, своей} & this is, why, people, people, very, tells, his \\
19 & \russian{украины, украине, российских, минобороны, российские, российской, подробности} & ukraine, ukraine, russian, defence, russian, russian, details \\
20 & \russian{время, который, словам, которая, которой, михаил, своего} & time, which, according to, which, which, Michael, his \\\\\\
    \end{tabular}
    }
    \caption{Highest probability words for each topic learned by 20-topic STM model}
    \label{tab:stm_topics}
\end{table*}

\clearpage

\section{MFC Data Preprocessing \& Classifier Training \& Label Generation}
\label{appendix:model-training}

\subsection{MFC Data Preprocessing}
We used MFC v4.0 \citep{card-etal-2015-media}, which contains MFC frame annotations of 27.7k articles across five social issues (death penalty, gun control, immigration, same-sex marriage, tobacco).
For each article, the data set provides annotations of two different granularity: article-level and span-level MFC frames.
Here, we chose to use span-level annotations to construct sentence-level labels as a sentence is a universal unit of texts; thus the trained models can be applied to both articles and comments in \dataname{} \citep{naderi-hirst-2017-classifying}. Also, training models based on large language models with article-level annotation could be challenging, as the cost of training exponentially grows as the input gets longer.

In order to convert the span-level labels in the MFC data to sentence-level, we first mapped each span annotation to its corresponding sentence\footnote{All articles were segmented into sentences using NLTK's sentence tokenizer.} in the article by finding a sentence that overlaps the most with each span. We then apply a rigorous filtering and kept the sentences and labels that at least two annotators agreed on. Since some articles have only one annotator, this process significantly reduces the number of data samples in the final training data\footnote{We conducted preliminary experiments with the more extensive, non-filtered data, but the trained models performed significantly worse.}. And for the cross-validation purpose, we randomly split the final training data into 10-folds and used eight folds among them as training, and one as development, and the last one fold as test set. 
In zero-shot classification scenarios, we held out the data from one social issue as test set and use the ramining data as training and development sets. \Tref{tab:mfc-data-stats} shows the final number of examples in different experiment settings.

\begin{table}[h]
\centering
\resizebox{\columnwidth}{!}{
\begin{tabular}{l|ccc}
\textbf{Experiment, Data}                        & \textbf{\# train} & \textbf{\# dev} & \textbf{\# test} \\\hline
In-domain, MFC unfiltered     & 2.1M     & 27K    & 27K     \\
In-domain, MFC                 & 105K     & 13K    & 13K     \\\hline
Zero-shot, MFC immigration                    & 100K     & 11K    & 21K     \\
Zero-shot, MFC same-sex              & 102K     & 11K    & 19K     \\
Zero-shot, MFC tobacco               & 106K     & 12K    & 13K    \\\hline
\end{tabular}
}
\caption{The number of training instances for each (Experiment set-up, test data) combination. All datasets consist of sentence-level MFC frame labels.}
\label{tab:mfc-data-stats}
\end{table}

\subsection{MFC Classifier Training}
\paragraph{Model}
Prior work has shown that classifiers based on large pre-trained language models achieve state-of-the-art performance. We built our MFC classifiers based on pre-trained language models in light of the findings. We specifically experimented with four different models: Roberta base (Roberta$_{B}$), Roberta large (Roberta$_{L}$), and XLM-R base (XLM-R$_{B}$), and XLM-R large (XLM-R$_{L}$) model.
Roberta$_{B}$ and XLM-R$_{B}$ both are the same size (125M parameters), and Roberta$_{L}$ and XLM-R$_{L}$ are also the same size (355M parameters).
We used 2-layer feedforward neural network, with the hidden size same as the base transformer model, for the final classifier layer.
The layer outputs 15-dimension logits to generate multi-class predictions over 15 frame classes.
We set the probability threshold for binary label conversion as 0.5. 

\paragraph{Hyperparameters}
The average length of sentences in the MFC sentence-level data is 37, with a standard deviation of 16. We set the maximum length of model inputs to 70 tokens, and truncated the inputs if they were longer than the limit. 
The learning rate and batch size were set to 5e-6 and 64, respectively. We used the AdamW optimizer with a schedule of  linearly decreasing the learning rate after the 5\% of warm-up steps. The model was trained with the cross-entropy loss for multi-class classification. The final model for each experiment was selected based on the macro F1 score over the development set. Each model was trained for 20 epochs, and in most cases, the best models were found before reaching ten epochs. 
In terms of training time, one epoch took less than 7 minutes for smaller models and 10 minutes for larger models when trained on a single GPU machine with A6000.
We did not do any hyperparameter search for our experiments, and performed exactly ten runs per experiment for 10-fold cross-validation. 

\paragraph{Evaluation Results} 
We computed F1 for each frame category following the standard multi-label evaluation procedure and used the macro averages across classes as our main metric.
\Tref{tab:res_framing_all_models} describes the evaluation results of four different baseline models and two different evaluation settings. 
The Roberta$_{L}$-based classifier achieves the best performance of 68.1.
One of few existing work we can compare against is \citet{naderi-hirst-2017-classifying}. They also converted the MFC data to sentence-level labels in a similar fashion and trained sentence-level MFC frame classifiers based on various LSTM models.
The best performance reported in the paper was 53.7 and our model unsurprisingly well surpasses the performance and establishes the new state-of-the-art.

The comparison across models over the in-domain MFC test set shows that multilingual models perform worse than monolingual models with the same number of parameters, which was also evidenced in  \citet{akyurek-etal-2020-multi}. It suggests that XLM-R enables seamless multilingual transfer at the cost of performance on English data. However, interestingly, the performance drop was lower with the larger multilingual model (XLM-R$_L$), suggesting that practitioners might want to consider large models when using multilingual models. 
\begin{table}[t]
\resizebox{\columnwidth}{!}{
\begin{tabular}{lllc}
                                                                        & \textbf{Data}                     & \textbf{Model}   & \textbf{Macro-F1}   \\\hline
\multirow{3}{*}{\begin{tabular}[c]{@{}l@{}}In-domain\end{tabular}} & \multirow{3}{*}{MFC} & Roberta$_{B}$ & 66.3$\pm$0.33 \\
                                                                        &                          & XLM-R$_{B}$   & 64.6$\pm$0.34     \\\cline{3-4}
                                                                        &                          & Roberta$_{L}$   & \textbf{68.1}$\pm$0.54     \\
                                                                        &                          & XLM-R$_{L}$  & 67.5$\pm$0.53      \\\hline
\multirow{4}{*}{Zero-shot}                                              & Immigration              & XLM-R$_{L}$ &  52.7$\pm$0.36    \\
                                                                        & Same-sex                  & XLM-R$_{L}$  &  50.4$\pm$0.64\\
                                                                        & Tobacco                  & XLM-R$_{L}$  & 51.0$\pm$0.33     \\\cline{2-4}
                                                                        & VoynaSlov            & XLM-R$_{L}$  &    \textbf{33.5}$\pm$0.72\\\hline 
\end{tabular}
}
\caption{Macro-F1 results of trained MFC classifiers. We conducted 10-fold cross-validation for each experiment and tested two scenarios: in-domain and zero-shot classification by leaving one social issue (Immigration, Same-sex, Tobacco) out of training data and using it as a test set. We also evaluate the classifiers with \dataname{}.}
\label{tab:res_framing_all_models}
\end{table}

\subsection{Generating MFC labels of \dataname}

\label{appendix:mfc-label-generation}

\citet{akyurek-etal-2020-multi} has examined several strategies of multilingual transfer of frame classifiers, and concluded that translating non-English target data to English and then applying the model (trained with only English data) over the translated text generally achieves the best performance. 
Following their suggestion, we translated all posts and comments in \dataname{} to English using the publicly available Russian-English machine translation model\footnote{We used the publicly released model, \href{https://huggingface.co/facebook/wmt19-ru-en}{facebook/wmt-ru-en}, downloaded from the HuggingFace model repository.}.
After translating, we apply the final model based on XLM-R$_{L}$ on each sentence in the translated English text and generate MFC frame labels.
Eventually, we are interested in the post-level and comment-level MFC frame labels of texts in \dataname, which might consist of more than one sentence.
We aggregated labels of sentences in a post/comment by majority voting with a random tie-breaking (i.e., hard voting) and one final frame label was assigned to each post/comment.

\section{Human Annotation of \dataname{}}
\label{appendix:mfc-annotation}

To measure how well the trained MFC frame classifiers work on \dataname{}, we annotated a small subset of Russian sentences in our data with the MFC frame labels. 
Although our trained classifiers generate sentence-level frame labels, in the end, we ultimately want to use the labels of articles/posts/comments for analyses.
We thus sampled the examples at the post-level, instead of individual sentence-level; We randomly sampled a post and added sentences in the post to the annotation set until we reached the desired number of sentences (50 each from state-affiliated and independent media). 
103 sentences from 49 articles were finally selected.
At the time of annotation, we provided the full post (i.e., all sentences in the post) along with the target sentence to annotate\footnote{We acknowledge that there is a discrepancy in amount of available information between our models and the human annotator. We believe the sentence-level annotations that consider the full context reflect what actual readers will perceive from reading the sentence more accurately. We leave incorporating global contexts for the sentence-level frame classifiers for future research.}.

\begin{figure*}[h]
    \centering
    \includegraphics[width=\linewidth]{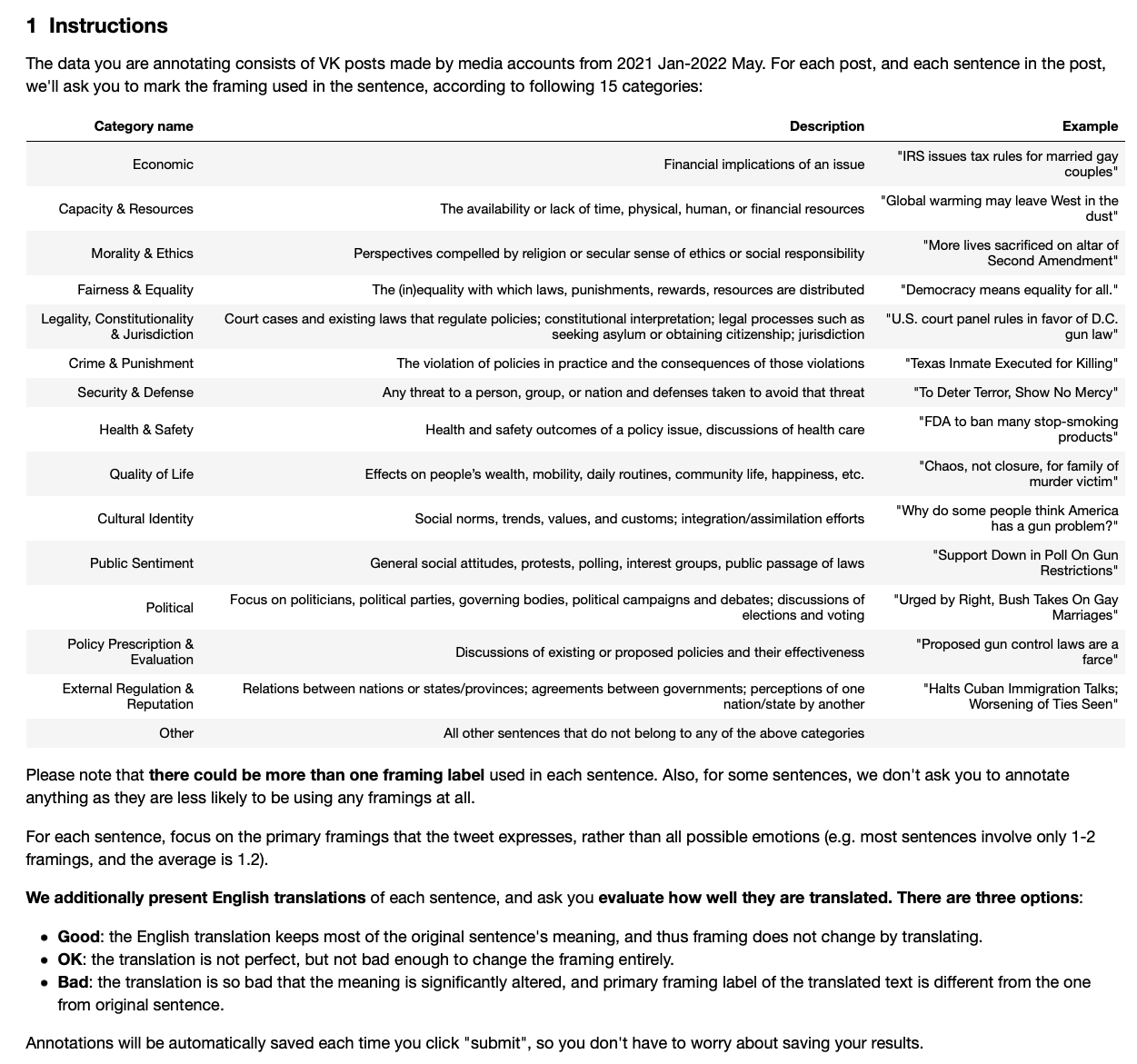}
    \caption{Screenshot of annotation instruction provided to the annotators. We borrowed the description from \citet{mendelsohn2021modeling} and the examples were selected from the sentence-level MFC training data we constructed.}
    \label{fig:annotation_instruction}
\end{figure*}

\begin{figure*}[h]
    \centering
    \includegraphics[width=\linewidth]{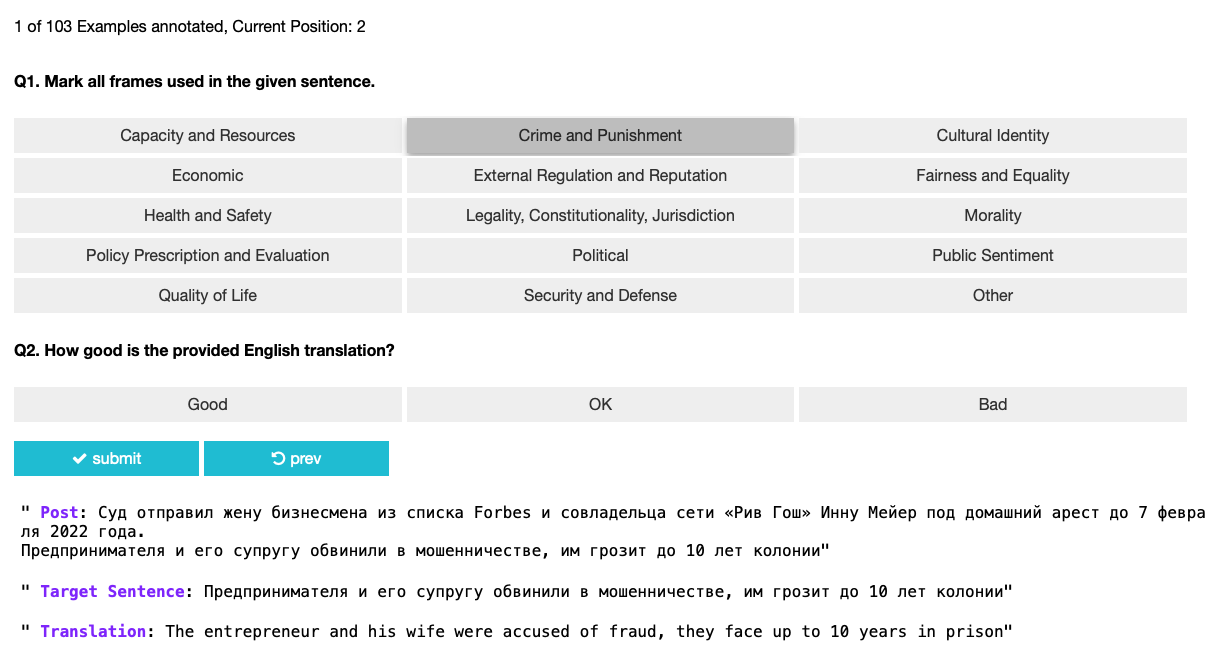}
    \caption{Screenshot of annotation user interface. For each target sentence, we provide the full context (\textit{Post}) and its English translation. Annotators mark their answers to two questions regarding the MFC frame and translation quality.}
    \label{fig:annotation_interface}
\end{figure*}

We recruited one native Russian speaker to annotate the sampled Russian sentences with the 15 MFC frame labels.
We ensured that the annotator have had past experiences annotating for similar tasks and has good understanding of the concept of framing. 
We acknowledge that the frame is inherently subjective and having more annotators could have resulted in a better quality evaluation set. However, we collected the in-domain annotation to get a broad sense of the generalizability.
During the annotation, we additionally presented the English translation of a target sentence (\Aref{appendix:mfc-label-generation}) and asked to annotate the quality of the translation to make sure the machine translation models are trustworthy and we can trust the inferred labels. 
The screenshots of the annotation instruction and the annotation interface are in \Fref{fig:annotation_instruction} and \Fref{fig:annotation_interface}.

We present label distribution of frame and translation quality labels of 103 annotated sentences in \Tref{tab:annotation-frame-state} and \Tref{tab:annotation-translation-state}. 
The annotation results over the MFC frames show that the MFC frame labels are imbalanced and so does the model performance. The final model performs especially poor on relatively abstract frames such as Capacity and Resources, Other, and Legality, Constitutionality.
On the other hand, the translation quality annotations suggest that the most of generated English translations of \dataname{} (99\%) are either good or decent enough to maintain the frame label in the original Russian text (i.e., OK). 
 
\begin{table}[t]
\resizebox{\columnwidth}{!}{
\begin{tabular}{lcc}
\textbf{Frame}                                     & \textbf{\#Label} & \textbf{F1}   \\\hline
Other                                     & 30           & 6.5  \\
Political                                 & 24           & 38.7 \\
Health and Safety                         & 14           & 62.9 \\
Crime and Punishment                      & 12           & 60.0 \\
Security and Defense                      & 11           & 54.6 \\
Legality, Constitutionality, Jurisdiction & 9            & 16.7 \\
External Regulation and Reputation        & 9            & 33.3 \\
Capacity and Resources                    & 6            & 0    \\
Quality of Life                           & 4            & 28.6 \\
Economic                                  & 2            & 33.3
\end{tabular}
}
\caption{The proportion of MFC frame labels in the annotated VoynaSlov data. F1 indicates the macro-F1 score of the final model measured for each frame class. The five frame classes that are not in this table did not attain any annotation label.}
\label{tab:annotation-frame-state}
\end{table}

\begin{table}[t]
\centering
\begin{tabular}{lc}
\textbf{Translation Quality} & \textbf{\#Label}\\\hline
Good & 59 \\
OK & 43 \\
Bad & 1 \\
\end{tabular}
\caption{The proportion of MFC frame labels in the annotated VoynaSlov data.}
\label{tab:annotation-translation-state}
\end{table}

\clearpage

\end{document}